\documentclass[10pt,twocolumn,letterpaper]{article}

\usepackage[pagenumbers]{cvpr} 

\usepackage{graphicx}
\usepackage{amsmath}
\usepackage{amssymb}
\usepackage{booktabs}
\usepackage[accsupp]{axessibility}
\usepackage{pifont}
\newcommand{\cmark}{\ding{51}}%
\newcommand{\xmark}{\ding{55}}%
\usepackage{multirow}
\usepackage{array}
\usepackage{arydshln}
\usepackage[pagebackref,breaklinks,colorlinks]{hyperref}
\usepackage{graphicx}

\usepackage[capitalize]{cleveref}
\crefname{section}{Sec.}{Secs.}
\Crefname{section}{Section}{Sections}
\Crefname{table}{Table}{Tables}
\crefname{table}{Tab.}{Tabs.}

\def\cvprPaperID{15028}
\def\confName{CVPR}
\def\confYear{2024}

\begin{document}

\title{Open-Vocabulary Attention Maps with Token Optimization for Semantic Segmentation in Diffusion Models}

\author{
Pablo Marcos-Manchón\textsuperscript{1, 2}\hfill
Roberto Alcover-Couso\textsuperscript{1}\hfill
Juan C. SanMiguel\textsuperscript{1}\hfill
José M. Martínez\textsuperscript{1}\\
{
\small{
\textsuperscript{1}VPULab, Autonomous University of Madrid, Spain
\hfill
\textsuperscript{2}Dynamics of Memory Formation Group, University of Barcelona, Spain
}}\\
{\tt\small pmarcos@ub.edu, \{roberto.alcover, juancarlos.sanmiguel, josem.martinez\}@uam.es}
}

\twocolumn[{%
\renewcommand\twocolumn[1][]{#1}%
\maketitle%
\centering%
\captionsetup{type=figure}%
\includegraphics[width=0.98\textwidth]{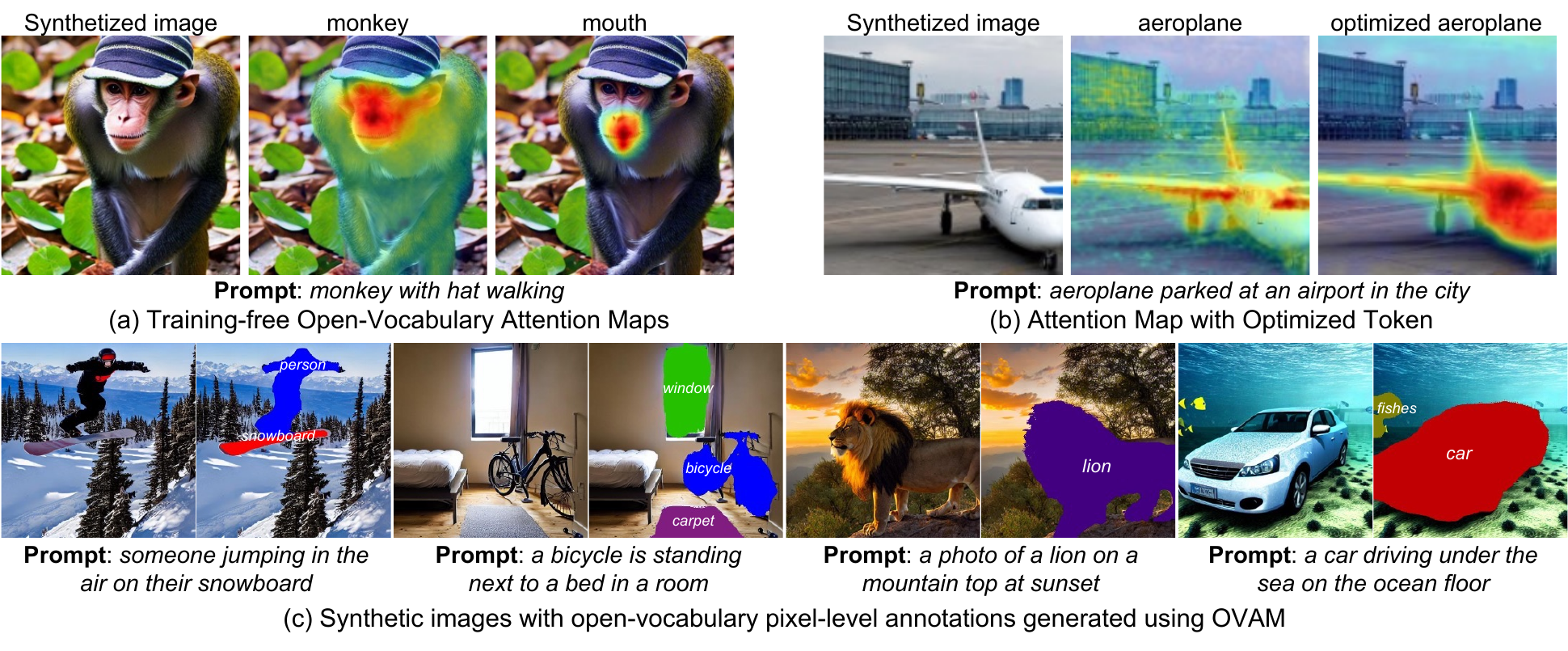}%
\captionof{figure}{
(a) We introduce Open-Vocabulary Attention Maps (OVAM), a training-free extension for text-to-image diffusion models to generate text-attribution maps based on open-vocabulary descriptions. Our approach overcomes the limitations of existing methods constrained by words contained within the prompt \cite{Attn2mask, DAAM, SecretSegmenter, diffumask}. (b) Our token optimization process enhances the creation of accurate attention maps, thereby improving the performance of existing semantic segmentation methods based on diffusion attentions \cite{Attn2mask, DAAM, DatasetDM, li2023grounded}. (c) Finally, we validate the utility of OVAM in producing synthetic images with precise pixel-level annotations.
}%
\label{fig:teaser}%
\vspace*{14pt}%
}]

\begin{abstract}

Diffusion models represent a new paradigm in text-to-image generation. Beyond generating high-quality images from text prompts, models such as Stable Diffusion have been successfully extended to the joint generation of semantic segmentation pseudo-masks. However, current extensions primarily rely on extracting attentions linked to prompt words used for image synthesis. This approach limits the generation of segmentation masks derived from word tokens not contained in the text prompt. In this work, we introduce Open-Vocabulary Attention Maps (OVAM)—a training-free method for text-to-image diffusion models that enables the generation of attention maps for any word. In addition, we propose a lightweight optimization process based on OVAM for finding tokens that generate accurate attention maps for an object class with a single annotation. We evaluate these tokens within existing state-of-the-art Stable Diffusion extensions. The best-performing model improves its mIoU from 52.1 to 86.6 for the synthetic images' pseudo-masks, demonstrating that our optimized tokens are an efficient way to improve the performance of existing methods without architectural changes or retraining. The implementation is available at \href{https://github.com/vpulab/ovam}{github.com/vpulab/ovam}.

\end{abstract}

\section{Introduction}
\label{sec:intro}

The introduction of diffusion models has led to a significant advancement in text-to-image (T2I) generation \cite{DiffusionBeatGans}. Diffusion-based models, such as Stable Diffusion \cite{sd} and other contemporary works \cite{Dalle2, sdxl, IMAGEN, kandinsky, wuerstchen, glide, instaflow}, have been rapidly adopted across the research community and industry, owing to their ability to generate high-quality images that accurately reflect the semantics of text prompts.

The image generation in diffusion models is driven by a denoising process, which iteratively refines noise from an initial noisy vector until a coherent image is synthesized \cite{DenoisingDiffusion, NonequilibriumThermodynamics}. To condition the image synthesis with a specific concept, usually represented by a text prompt, models utilize cross-attention mechanisms throughout the denoising steps \cite{sd, AttentionIsAllYouNeed}. These mechanisms yield cross-attention matrices that facilitate the incorporation of semantic details into the spatial layout of images.
Due to their role in fusing spatial and semantic information, these matrices have become a key part of works for interpreting the text prompt's influence on image layout \cite{DAAM} and for developing methods that extract pixel-level semantic annotations from the diffusion process \cite{diffumask, DatasetDM, Attn2mask, li2023grounded, SecretSegmenter, ODISE, DatasetDiffusion, xiao2023text}.

The use of attention matrices to relate semantic information to spatial layout draws inspiration from natural language processing, where word-specific attention has been shown to correlate with lexical attribution \cite{BertLookAt, AttentionNotNot}. In the context of T2I diffusion models, extracting attention matrices during image generation has become the primary method for extending the models' ability to jointly synthesize images and generate semantic segmentation pseudo-masks \cite{DatasetDM, Attn2mask, li2023grounded, SecretSegmenter, DatasetDiffusion}. This method appears to be a promising approach to addressing the data scarcity challenge in semantic segmentation training, a problem arising from the high costs of pixel-level semantic annotation \cite{Lin_2019_ICCV}.

Existing methods that directly utilize attention matrices for generating semantic segmentation are limited by text prompt tokens, requiring an association between each semantic class and a prompt word \cite{li2023grounded, Attn2mask, DAAM, SecretSegmenter}. However, as shown in Fig. \ref{fig:teaser}, not all object classes are explicitly mentioned in the text prompts, which highly limits the flexibility of these methods. To address this issue, some strategies incorporate additional modules that employ these attention features for mask generation; however, this necessitates additional supervised training, thereby limiting the methods' domain and increasing computational costs \cite{DatasetDM, diffumask}. 

In response to the abovementioned challenges, we introduce Open-Vocabulary Attention Maps (OVAM), a training-free approach that generalizes the use of attention maps from image synthesis. OVAM enables the creation of semantic segmentation masks described by an open vocabulary, irrespective of the words in the text prompts used for image generation.
Moreover, we propose a token optimization process based on OVAM, which allows for the learning of open-vocabulary tokens that generate accurate attention maps for segmenting an object class with just a single annotation per class (see Fig. \ref{fig:teaser}). These tokens not only enhance the quality of segmentation masks produced by OVAM but also, as we demonstrate through our experiments, they can improve the performance of existing open-vocabulary segmentation methods without any modifications or additional training \cite{DatasetDM, DAAM, Attn2mask, li2023grounded}.

\section{Related work}
\label{sec:related-work}

\textbf{T2I Diffusion Models.} 
The release of pretrained T2I diffusion models \cite{sd, Dalle2, sdxl, IMAGEN, kandinsky, wuerstchen, glide, instaflow, diffusionSurvey} has advanced the application of image generation in both research and practical contexts. A key to their success has been the decomposition of the synthesis process into sub-parts: a projection of input text \cite{clip, t5}, diffusion through a denoising network \cite{DenoisingDiffusion, NonequilibriumThermodynamics}, and the decoding of final images. This modularity has facilitated the adaptation of these models to new tasks. Our method is centered on adapting Stable Diffusion \cite{sd} for semantic segmentation, although it is general enough to be applied to similar models that incorporate attention mechanisms within the denoising network \cite{AttentionIsAllYouNeed}.

\textbf{Use of Attention in Diffusion Models.} 
Attention mechanisms have a key role in the control of the image generation process, making attention matrices a central component for adapting diffusion models to various tasks. These matrices are an essential element in works pertaining to image editing \cite{AttendAndExcite, PromptToPrompt, imageTranslation, zhang2023realworld, Tumanyan_2023_CVPR}, adding layout constraints \cite{T2IAdapter, Couairon_2023_ICCV}, model interpretability \cite{DAAM, park2023understanding}, semantic correspondence \cite{hedlin2023unsupervised, luo2023dhf, tang2023dift, zhang2023tale}, and diverse segmentation tasks\cite{ODISE, DatasetDM, SecretSegmenter, diffuseAttendSegment, Attn2mask, li2023grounded, diffumask, DatasetDiffusion, PNVR_2023_ICCV, ni2023refdiff, xiao2023text}. Our work employs those attentions to generate semantic segmentation ground truth, extending diffusion models to generate synthetic images with corresponding semantic segmentation pseudo-labels.

\textbf{Open-Vocabulary Segmentation.} 
Open-vocabulary semantic segmentation aims to divide an image into regions based on textual descriptions \cite{wang2023ovvg}. State-of-the-art proposals mostly rely on pretrained language models \cite{ovseg, lsegplus, freeseg, ding2021decoupling, xu2021}, and recent approaches focus on T2I pretrained diffusion models. Diffusion-based systems range from training-free methods \cite{Attn2mask, SecretSegmenter, DAAM, DatasetDiffusion, PNVR_2023_ICCV} to those incorporating additional trained modules that utilize diffusion features \cite{DatasetDM, diffumask, li2023grounded, ODISE, ni2023refdiff, ovdiff}. Our approach introduces a training-free methodology using pretrained diffusion models, overcoming the limitations of prior training-free approaches by generating semantic segmentation masks independent of the text prompt's vocabulary constraints.

\textbf{Token Optimization in Diffusion Models.} 
Token optimization involves refining input text embeddings used for image generation  while the rest of the diffusion model remains unchanged. This technique has been employed for various objectives, such as generating synthetic images imitating a target image \cite{hardprompts, Schwartz_2023_ICCV} or learning new concepts from a limited number of examples \cite{KeyLockedRank, textualInversion, breakaSCENE}. Our research applies an optimization strategy to train text-embedding tokens for class segmentation. These tokens are then used to improve the accuracy of segmentation masks created by OVAM and existing diffusion-based segmentation methods without necessitating modifications to their architecture or additional training \cite{DatasetDM, li2023grounded, DAAM, Attn2mask}.

\begin{figure*}[ht]
    \centering
    \includegraphics[width=0.95\textwidth]{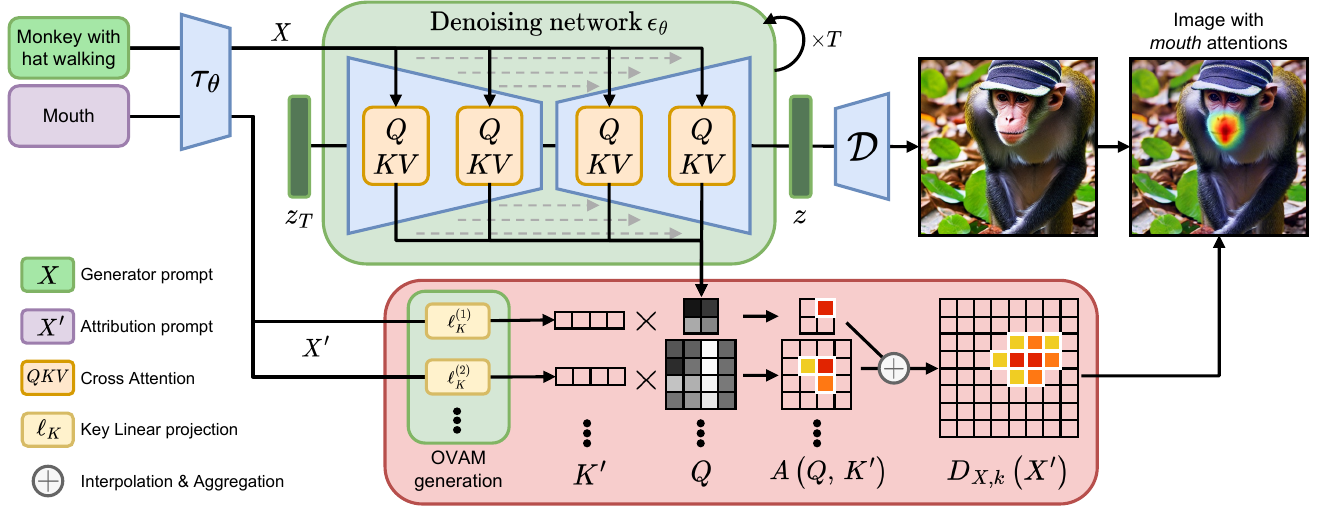}
    \caption[OVAM construction diagram]{A schematic representation of the OVAM generation process (red module) utilizing the Stable Diffusion architecture \cite{sd} (rest of modules). The example synthesizes an image using the generator prompt \emph{monkey with hat walking}. During the OVAM generation, pixel queries \(Q\) are extracted from the denoising network. These pixel queries are combined with the text embedding $K'$ corresponding to the attribution prompt \emph{mouth}, constructing the OVAM heatmap $D_{X,k}(X')$, which highlights the monkey's mouth in the synthesized image.}
    \label{fig:diagram-ovam}
\end{figure*}

\section{Proposed methodology}
\label{sec:proposed-work}

\subsection{Cross-Attention Formulation}
\label{subsec:formulation}

The denoising process, applied during image synthesis in diffusion models, takes place in a lower-dimensional latent space with shape \(W \times H \times C\). Within this space, a latent vector—\(z_T\)—is transformed by the denoising network until they obtain a latent representation of the final image \cite{DenoisingDiffusion}.

In Stable Diffusion \cite{sd}, the UNet used as the denoising network contains convolutional downsampling and upsampling blocks \cite{UNET}. These blocks process the latent vector \(z_t\) at each time step, iteratively generating the image. 

At each time step \(t\), the \(i\)-th convolutional block outputs a vector, denoted as \(h_{i,t} \in \mathbb{R}^{\lceil \frac{W}{r^{(i)}} \rceil \times \lceil  \frac{H}{r^{(i)}} \rceil \times C^{(i)} }\), where \(r^{(i)}\) is a reduction factor tied to the block's resolution. These vectors contain the spatial information of the process. A text encoder, \(\tau_\theta\), is used to convert the input text prompt into a text embedding \(X \in \mathbb{R}^{l_E \times l_X}\) that consists of \(l_X\) tokens, each of dimension \(l_E\). This embedding, which contains the semantic information, is subsequently combined using multi-headed cross-attention layers at each block's output.

Each cross-attention takes three inputs: a query \(Q\), a key \(K\), and a value \(V\) \cite{AttentionIsAllYouNeed}. While \(K\) and \(V\) are computed from linear projections of the text embedding \(X\), the input \(Q\) is sourced from a projection of the convolutional blocks' outputs. These linear projections, denoted as \(\ell^{(i)}\), project $X$ and $h_{i,t}$ into \(l_H^{(i)}\) attention heads to build $Q$, $K$ and $V$:

\begin{equation}\label{eqn:linear-proyection}
      Q_{i, t} = \ell_{Q}^{(i)} \left(h_{i, t}\right), \,K_{i}=\ell_{K}^{(i)}\left(X\right), \, V_{i} = \ell_{V}^{(i)}\left(X\right)\,.
\end{equation}

The components Q, K, V are combined in the cross-attention blocks. As a result, during the process a cross-attention matrix \(A\left(Q_{i,t}, \, K_{i}\right)\in \mathbb{R}^{\lceil \frac{W}{r^{(i)}} \rceil \times \lceil  \frac{H}{r^{(i)}} \rceil \times l_H^{(i)} \times l_X}\), is computed for each block $i$ and time step $t$:

\begin{gather}
\label{eqn:cross-attention}
    \text{CrossAttention}(Q_{i, t},\,  K_{i}, \,V_{i}) = A\left(Q_{i, t}, \, K_{i}\right)\cdot V_{i}  \, ,  \\
    A\left(Q_{i,t}, \, K_{i}\right) = \text{softmax}\left ( \frac{Q_{i, t} K_{i}^T}{\sqrt{d}}\right ) \, .
\end{gather}

These cross-attention matrices weigh the influence of each token from $X$ on the image's pixels, establishing a direct correlation between the spatial layout and the semantic content of the text. Studies like DAAM \cite{DAAM} utilize these token-aggregated matrices to discern the influence of each prompt word on the resulting image. Moreover, these matrices are employed as input features for open-vocabulary semantic segmentation systems based on diffusion models \cite{diffumask, SecretSegmenter, Attn2mask, li2023grounded, ODISE, DatasetDM}. However, direct extraction restricts open-vocabulary mask generation to the tokens within $X$. To overcome this limitation, our approach introduces Open-Vocabulary Attention Maps (OVAM), which generalize these matrices and eliminate this constraint.

\subsection{Open-Vocabulary Attention Maps}

For the construction of Open-Vocabulary Attention Maps (OVAM), we introduce a second text prompt, which we refer to as the \emph{attribution prompt}. This text allows us to control the attention heatmaps using open vocabulary (see Fig. \ref{fig:diagram-ovam}), eliminating the constraint of using the text prompt employed for image generation. The diffusion model's text encoder \(\tau_\theta\) is employed to transform the attribution prompt to produce an associated text embedding \(X^\prime \in \mathbb{R}^{l_E \times l_{X^\prime}}\). This embedding consists of \(l_{X^\prime}\) tokens, which may not match the length of \(X\),  with dimension \(l_E\).
For generating the attention matrices, we project $X'$ using the key projections \( \ell_{K}^{(i)}\) to generate attribution keys $K'$:

\begin{equation}\label{eqn:attribution-key}
    K^\prime_{i} = \ell_{K}^{(i)} \left ( X^\prime \right ) \,.
\end{equation}

These attribution keys \(K^\prime\) are combined with the pixel queries \(Q_{i, t}\) computed during the denoising process (refer to Eqn. \ref{eqn:linear-proyection}), creating the open-vocabulary attention matrices as \(A\left(Q_{i,t}, K_{i}^\prime\right)\) (see Eqn. \ref{eqn:cross-attention}). These matrices, which capture the influence of tokens on the image, have dimensions \(\lceil \frac{W}{r^{(i)}} \rceil \times \lceil  \frac{H}{r^{(i)}} \rceil \times l_H^{(i)} \times l_{X^\prime}\), including two spatial dimensions, the attention heads, and the tokens of \(X^\prime\). To generate the OVAM corresponding to the $k$-th token of $X'$, we aggregate the matrices across blocks, timestamps, and attention heads, following the same procedure used in \cite{DAAM}:

\begin{equation}\label{eqn:ovam-aggregation}
\small
    D_{X, k} \left ( X^\prime\right ) = \sum_{i, t, h} \text{resize} \left ( A_{h, k} \left( Q_{i, t}, \, K^\prime_{i}\right)\right ) \, \in \mathbb{R}^{W \times H},
\end{equation}
where the notation \(A_{h, k}\) refers to the slice associated with the \(h\)-th attention head and the \(k\)-th token.  For matrices of varying resolutions, bilinear interpolation is used for resizing to a common resolution. Figure \ref{fig:diagram-ovam} shows the construction process of these maps for text attribution.

This approach can be viewed as a generalization of the DAAM method \cite{DAAM} and others that directly extract cross-attentions to generate semantic segmentation masks \cite{Attn2mask, diffumask, SecretSegmenter}, offering a more versatile framework.
When both \(X\) and \(X^\prime\) are identical, the heatmaps \(D_{X, k} \left ( X^\prime\right )\) are equivalent to directly extracting and aggregating the cross-attention matrices computed during image synthesis.

\subsection{Token Optimization via OVAM}
\label{subsec:token-optimization}
\begin{figure}[t]
    \centering
    \includegraphics[width=0.75\columnwidth]{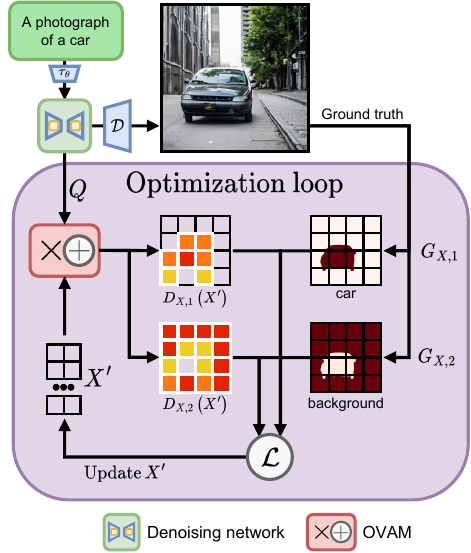}
    \caption{Diagram illustrating the optimization process for an $X'$ composed of two tokens: a \emph{background} and a \emph{car} token. OVAM generates one heatmap for each token, and the optimization updates $X'$ to align the attentions generated with the target mask.}
    \label{fig:monkey-optimization}
\end{figure}

The proposed OVAM (Eqn. \ref{eqn:ovam-aggregation}) relies on spatial information from cross-attention queries \(Q_{i, t}\), derived from \(X\) and the initial state of the diffusion process, alongside semantic information from keys \(K_{i}^\prime\), computed from \(X^\prime\). When \(Q_{i, t}\) is fixed, OVAM acts as a mapping from \(X^\prime\) to attribution maps for each token it contains. One challenge in open-vocabulary segmentation lies in selecting the most effective descriptors. For example, is ``mouth'' the best descriptor for the monkey's mouth (see Figure \ref{fig:diagram-ovam})?. While some open-vocabulary segmentation studies address this by averaging synonyms and using prompts in varied contexts \cite{ovseg}, we formulate descriptor selection as an optimization over \(X^\prime\).

To perform the optimization, we define \(X'\) with one token for each class we aim to optimize. Initially, $X'$ is initialized with tokens corresponding to the class names for each segmentation class we target, including a background class initialized with \textlangle\emph{SoT}\textrangle, known to effectively capture background information in Stable Diffusion \cite{DAAM}. Additionally, we require ground truth \(G_X \in \mathbb{R}^{W \times H \times l_{X^\prime}}\) for an image synthesized from prompt \(X\). This ground truth is a semantic segmentation map comprising \(l_{X^\prime}\) classes, \(G_{X, k}\), with each class corresponding to a token of \(X'\).

For the general case with multiple images, consider a set \(\left \{ \left (X_j, G_{X_j} \right ) \right \}_{j}\), where each \(X_j\) is a prompt embedding used for image synthesis, and \(G_{X_j}\) is the corresponding ground truth. Our goal is to jointly optimize the tokens of \(X^\prime\) to segment images consistent with the ground truth:
\begin{equation}\label{eqn:optimization}
    X^* = \underset{X^\prime}{\text{argmin}} \, \sum_{j} \mathcal{L} \left ( D_{X_{j}} \left (X^\prime\right), \, G_{X_j} \right ) \,,
\end{equation}
where \(\mathcal{L}\) is a loss function measuring the discrepancy between the OVAM heatmap and the ground truth mask. We employ the binary cross-entropy \cite{lossesSurvey} as training loss:

\begin{equation}\label{eqn:optimization-loss}
\mathcal{L} \left ( D_{X_{j}} \left (X^\prime\right), \, G_{X_j} \right ) =
     \sum_{k=1}^{l_{X^\prime}} \text{BCE} \left ( D_{X_{j}, k} \left (X^\prime\right), \, G_{X_j, k} \right ).
\end{equation}

\begin{figure}
    \centering
    \includegraphics[width=\columnwidth]{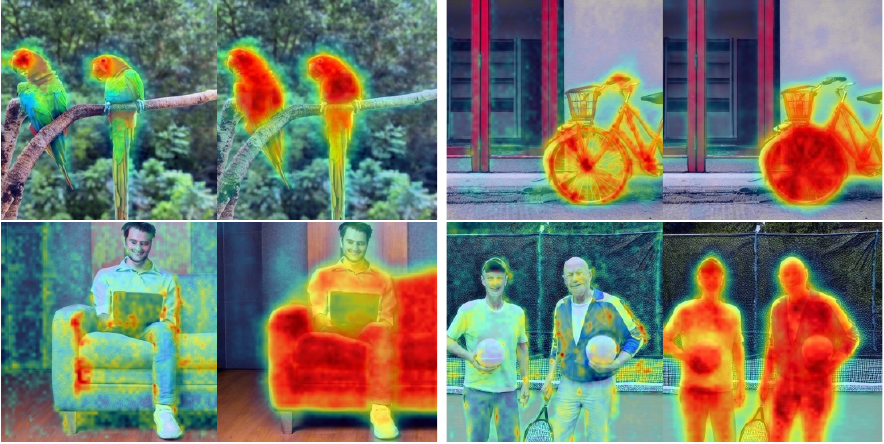}
    \caption{Comparative visualization of attention maps. Left images show attention using class name tokens (\emph{bird}, \emph{bicycle}, \emph{sofa} and \emph{person}) while the images on the right use optimized tokens with a training set that does not contain these images.}
    \label{fig:example-optimized}
\end{figure}

Given that the objective is differentiable with respect to \(X^\prime\), gradient descent is used for optimization. This method is similar to the conventional training of semantic segmentation models but computationally efficient since it involves learning a reduced number of parameters and can be completed in less than a minute on a single GPU. Figure \ref{fig:monkey-optimization} illustrates the optimization process, and Figure \ref{fig:example-optimized} shows examples of attention maps from optimized tokens. The experimental section further evaluates the efficacy of these tokens in open-vocabulary segmentation and their adaptability to different systems without requiring architectural changes.

\subsection{Mask Binarization}

To generate binary segmentation masks from OVAM, we apply a refinement based on self-attentions and binarize the continuous heatmaps using a fixed threshold.

The denoising network of Stable Diffusion \cite{sd}, besides integrating cross-attention mechanisms for merging spatial and semantic information, also includes self-attention blocks \cite{AttentionIsAllYouNeed}. Unlike cross-attention, self-attention mechanisms do not directly relate semantics with spatial layout but capture object groupings information. For this reason, they are utilized to generate and refine segmentation masks in diffusion-based studies \cite{diffuseAttendSegment, DatasetDiffusion, SecretSegmenter, xiao2023text}. Since we seek the most granular information to refine the masks, we exclusively extract self-attention matrices from the highest-resolution blocks, $W \times H$, and aggregate them across blocks, heads, and time steps to produce a fused map:

\begin{equation}
\label{eqn:self-attention}
    \mathcal{A}_\alpha = \underset{[\alpha, 1]}{norm} \left (\sum_{i, h, t} A_{t, i, h}^{self}\right ) \in \mathbb{R}^{W \times H} \,.
\end{equation}

To combine these self-attention maps with OVAM heatmaps, we first normalize their values using min-max scaling, setting the range between $\alpha$ and 1. This normalization allows us to fuse the self-attention map with the OVAM heatmap multipliying both maps. The hyperparameter $\alpha$ allow the control of self-attention's impact on the final mask. Subsequently, we apply a threshold binarization relative to the peak value of the combined heatmap:

\begin{equation}
\label{eqn:binarization}
    D_{X, k}^{\mathbb{I}_{\alpha,\tau}} \left ( X' \right ) = \mathbb{I} \left ( \mathcal{A}_{\alpha} \cdot D_{X, k} (X')\ge \tau M \right ) \, ,
\end{equation}
where $M=\max \mathcal{A}_{\alpha} \cdot D_{X, k} (X')$.

Finally, these binary masks are refined using dense Conditional Random Fields \cite{crf}. This process uses the image's geometry to improve the masks, adjusting their alignment with objects and improving the accuracy of the semantic segmentation.

\section{Experimental Results}
\label{sec:experiments}

\subsection{Evaluating Generated Pseudo-Masks}
\label{subsection:exp-1}

First, we perform an experiment to assess the quality of OVAM-generated pseudo-masks and compare them with those extracted by other Stable Diffusion-based methods.

\textbf{Dataset Generation}. Two datasets were produced using varying prompting strategies to compare the quality of generated pseudo-masks. Initially, a reduced dataset named \emph{VOC-sim} was created to measure mask quality in a simplified context, partially replicating experiments from \cite{li2023grounded}. This dataset was generated using Stable Diffusion \cite{sd} with prompt templates of the form \emph{A photograph of a \textlangle classname\textrangle}, utilizing the 20 VOC classes \cite{pascal-voc-2012} to create a total of 600 images, with an equal number of images per class. To generate different images using the same prompt, the diffusion model's seed was varied. Subsequently, a more complex dataset, \emph{COCO-cap}, was generated. It was based on 1,100 text captions sampled from COCO \cite{cocoCaptions} containing one of the VOC classes objects. These richer text prompts were used to generate images with complex scenes.

\textbf{Token Optimization}. For each class, we optimized a single token using only one image using the prompt \emph{A photograph of a \textlangle classname\textrangle}, employing distinct seeds from those for the evaluation set. We conducted separate optimizations for each class with \(X'\) consisting of two tokens, representing \emph{background} and the \emph{class} (refer to Fig. \ref{fig:monkey-optimization}). The optimization was performed over 500 epochs on an A40 GPU, with the process for each token taking less than a minute. These tokens, corresponding to the 20 VOC classes, were used in all further experiments.

\textbf{Masks Generation with OVAM}. When generating synthetic images using Stable Diffusion 1.5, we produced OVAM binary masks, with both optimized and non-optimized variants, for comparison. For the non-optimized evaluation, we utilized the attribution prompt \(X'=\text{\emph{A photograph of \textlangle classname\textrangle}}\), extracting the heatmap associated with the classname. This approach was applied in the \emph{VOC-sim} evaluation, where it matches the text prompt, as well as in \emph{COCO-cap}, where the prompts differ. For binarization, we empirically determined the hyperparameters \(\tau=0.4\) (also used in \cite{DAAM}) and \(\alpha=0.85\) (Eqn. \ref{eqn:binarization}), which were suitable choices in initial experiments with preliminary datasets. In evaluations with optimized tokens, we used the attribution prompts \(X'\) obtained from the optimization process (see Fig. \ref{fig:monkey-optimization}), setting parameters to \(\tau=0.8\) and \(\alpha=0.95\). We use the dCRF \cite{crf} implementation \emph{SimpleCRF} \cite{dCRFImplementation} with the default parameters.

 \textbf{Mask Generation with Other Methods}. To benchmark OVAM's performance, we replicate the experimental protocol with the same prompts and seeds using various Stable Diffusion-based pseudo-mask generation methods. We compared OVAM with the training-free methods DAAM \cite{DAAM} and Attn2Mask \cite{Attn2mask}, which generate segmentation masks using cross-attentions. Due to the absence of a public implementation, we recreated Attn2Mask based on details from the original work. In the \emph{COCO-cap} dataset, the class name does not always appear in the prompt; sometimes, a synonym is used or inferred from the text context. Therefore, these methods, which generate masks using cross-attention linked to prompt words, cannot create masks for all images in \emph{COCO-cap}. To address this, we modified them to employ Open-Vocabulary Attention Maps, yielding identical results when the word is in the text prompt and enabling generation across all images. 
 Furthermore, we assessed the masks created by Grounded Diffusion \cite{li2023grounded} and DatasetDM \cite{DatasetDM}. As these methods incorporate additional trained modules for mask generation, we employ the pre-trained weights provided by the authors for creating masks for VOC classes.

\textbf{Evaluation}. The pseudo-masks were manually annotated for the evaluation. Due to different parameters of Stable Diffusion required for some methods (e.g. the number of timesteps), synthesized images contain small variations. Nevertheless, we annotated each unique image, including those with subtle variations, to ensure a comparable evaluation. Three annotators were tasked with labeling the primary object in each image, excluding any without a clear object. We evaluated the mask of this main object against the corresponding automatically generated pseudo-masks using \( \text{mIoU} = \frac{1}{20} \sum_{c} \text{IoU}_c \), where \( \text{IoU}_c = \frac{TP_c}{TP_c + FP_c + FN_c} \).
\vspace{-2mm}

\textbf{Results}. Table \ref{table:ov-grounding} summarizes the results. We observe that training-free methods achieve comparable outcomes, with variations likely due to distinct processing applied to attention maps. OVAM with token optimization outperforms models that require additional training, even though it relies on only a single annotation per class. The performance of Grounded Diffusion is noteworthy; however, it is skewed due to lower performance in several classes, which is addressed in the subsequent experiment.

\begin{table}[ht]
\centering
\setlength{\tabcolsep}{4pt} 
\resizebox{\columnwidth}{!}{ 
\begin{tabular}{l|c||c|c}
\hline
\multirow{2}{*}{Method} & \multirow{2}{*}{Vocab.} & \multicolumn{2}{c}{Dataset (mIoU \%)} \\ 
\cline{3-4}
 &  & VOC-sim & COCO-cap \\ 
\hline
\multicolumn{4}{l}{\emph{Training-free methods}} \\
DAAM \cite{DAAM} & Prompt & 66.2 & 48.4 \\ 
Attn2Mask \cite{Attn2mask} & Prompt & 68.7 & 55.0 \\
OVAM (w/o token opt.) (ours) & Open & \textbf{70.4} & \textbf{58.2} \\ \hline
\multicolumn{4}{l}{\emph{Methods with additional training}} \\
Grounded Diffusion \cite{li2023grounded} & Open & 62.1 & 50.2 \\
DatasetDM \cite{DatasetDM} & Open & 80.3 & 59.3 \\ 
OVAM + token opt. (ours) & Open & \textbf{82.5} & \textbf{69.2} \\  \hline
\end{tabular}
}

\caption{Comparing pseudo-mask generation performance on \emph{VOC-sim} and \emph{COCO-cap}. For \emph{COCO-cap}, methods bound by prompt vocabulary have been adapted to employ attention from attribution prompts for evaluating instances where the class name is not explicitly present in the prompts.}

\label{table:ov-grounding}
\end{table}

\subsection{Token Optimization with Different Methods}

\begin{figure}
    \centering
    \includegraphics[width=0.88\columnwidth]{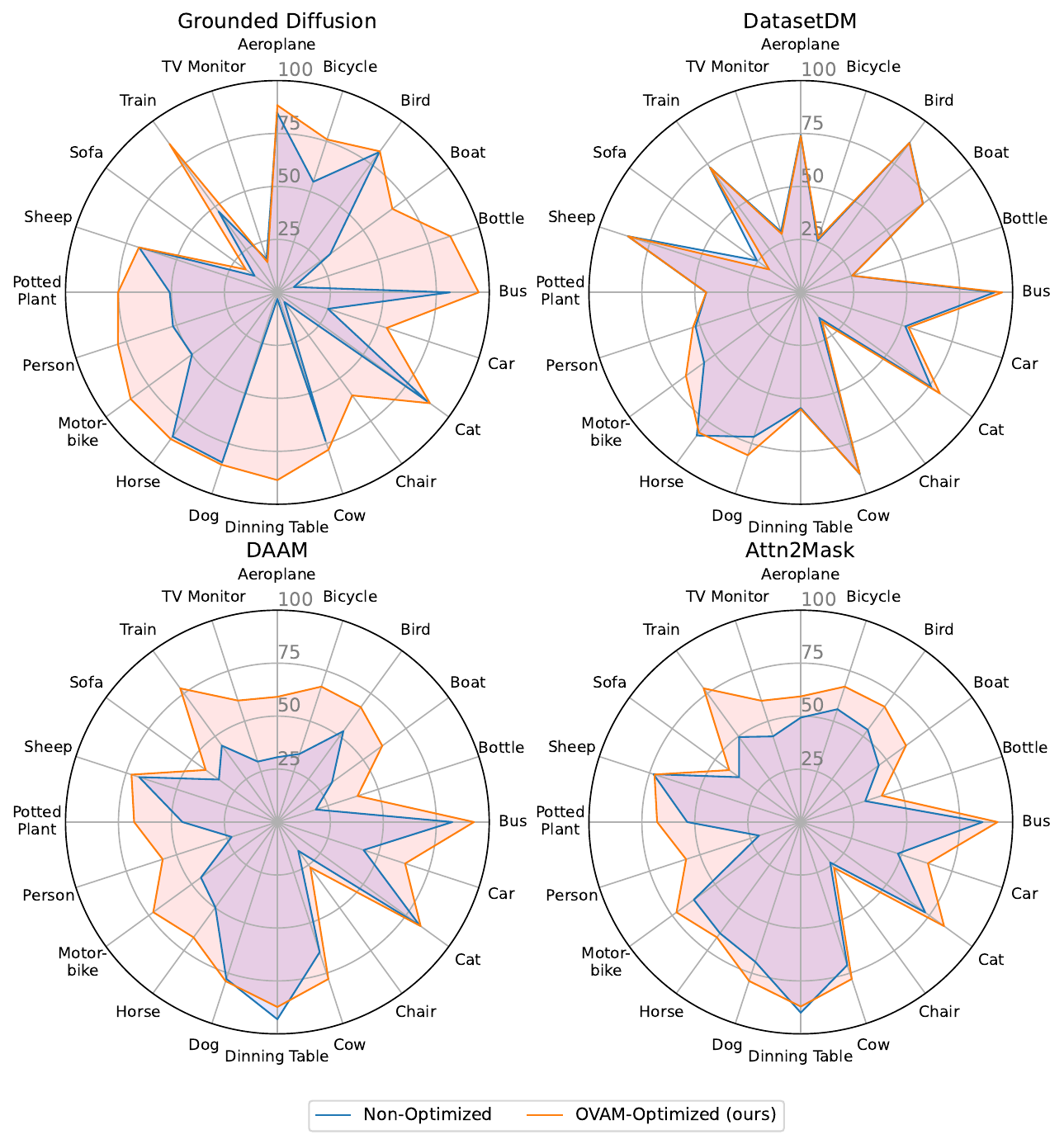}
    \caption{Class-performance comparison (IoU) of pseudo-masks generated by existing methods with/without the OVAM's token optimization, for the classes of the synthetic dataset \emph{COCO-cap.}}
    \label{fig:optimization-radar}
    \vspace{-3mm}
\end{figure}
\textbf{Use of Token Optimization in Other Methods}. We replicated the evaluation of the prior experiment, applying token optimization across all techniques. Methods with additional training  \cite{li2023grounded, DatasetDM}, contains modules capable of evaluate open-vocabulary tokens for mask description. Instead of using the name of the class, as used in original papers, we utilized OVAM-optimized tokens. This integration is effective because all the methods are based on the cross-attention mechanisms of Stable Diffusion. For the training-free methods \cite{DAAM, Attn2mask}, we employed the adaptations compatible with open-vocabulary attention maps, which allow for the utilization of any arbitrary tokens to describe the masks.

\textbf{Evaluation}. Like changing the token used as descriptor of the pseudo-mask does not change the image generation, we repeat the evaluation performed in the previous experiment using \emph{VOC-sim} and \emph{COCO-cap} synthetic datasets.

\textbf{Results}. Figure \ref{fig:optimization-radar} displays the results for all classes, and Table \ref{tab:token-opt-comparison} offers detailed outcomes for selected classes. Token optimization has improved the performance across all methods tested, improving the mIoU on all classes in methods based exclusively on cross-attentions \cite{DAAM, Attn2mask, li2023grounded}. DatasetDM \cite{DatasetDM} shows a minor improvement, as its mask generation is partly based on additional features from the diffusion process. Grounded Diffusion \cite{li2023grounded} showed notable improvement, likely due to low performance caused by misaligment between class names and the target concept on some classes, being corrected by the more precise class descriptions provided by the optimized tokens.

\begin{table*}[ht]
\centering
\setlength{\tabcolsep}{3.1pt} 
\small 
\resizebox{\linewidth}{!}{%
\begin{tabular}{l||cccccccccc|c|c}
\hline
\multirow{2}{*}{Method} & \multicolumn{10}{c|}{Selected classes (COCO-cap IoU \%)} & \multicolumn{2}{c}{Dataset (mIoU \%)} \\
\cline{2-13}
 & aeroplane & bicycle & boat & bus & car & cat & cow & dog & motorbike & person & VOC-sim & COCO-cap \\
\hline
\multicolumn{13}{l}{\emph{Training-free methods}}  \\
DAAM \cite{DAAM} & 30.6 & 33.8 & 31.9 & 82.6 & 42.8 & 83.0 & 64.6 & 77.9 & 44.6 & 22.7 & 66.2 & 48.4 \\
DAAM + \emph{token optimization} & \textbf{59.1} & \textbf{67.2} & \textbf{61.2} & \textbf{92.8} & \textbf{63.4} & \textbf{83.6} & \textbf{77.9} & \textbf{79.0} & \textbf{72.4} & \textbf{56.9} & \textbf{79.7} & \textbf{66.1} \\  \hdashline
Attn2Mask \cite{Attn2mask} & 49.3 & 56.0 & 45.5 & 85.8 & 48.2 & 72.6 & 71.0 & 69.4 & 62.4 & 20.7 & 68.7 & 55.0 \\
Attn2Mask +  \emph{token opt.} & \textbf{59.3} & \textbf{67.2} & \textbf{61.4} & \textbf{92.9} & \textbf{63.1} & \textbf{83.6} & \textbf{77.9} & \textbf{78.9} & \textbf{72.5} & \textbf{56.9} & \textbf{81.9} & \textbf{66.1} \\  \hdashline
OVAM (ours) & 65.1 & 64.3 & 51.9 & 84.9 & 47.5 & 67.9 & 76.5 & 65.8 & 69.4 & 19.7 & 70.4 & 58.2 \\
OVAM +  \emph{token optimization} & \textbf{67.8} & \textbf{68.4} & \textbf{64.6} & \textbf{94.5} & \textbf{63.2} & \textbf{87.6} & \textbf{82.4} & \textbf{81.9} & \textbf{74.2} & \textbf{60.9} & \textbf{82.5} & \textbf{69.2} \\  \hline
\multicolumn{13}{l}{\emph{Methods with additional training}}  \\
DatasetDM \cite{DatasetDM} & \textbf{74.1} & 25.7 & \textbf{71.3} & 91.4 & 51.9 & 76.2 & \textbf{90.1} & 71.7 & 56.4 & 52.2 & 80.3 & 59.3 \\
DatasetDM + \emph{token opt.} & 73.7 & \textbf{26.7} & 71.2 & \textbf{95.1} & \textbf{53.5} & \textbf{81.2} & 89.9 & \textbf{80.8} & \textbf{67.0} & \textbf{53.5} & \textbf{80.6} & \textbf{60.5} \\ \hdashline
Grounded Diffusion \cite{li2023grounded} & 84.6 & 54.9 & 30.8 & 81.4 & 25.1 & 87.3 & 73.7 & 84.4 & 49.8 & 51.8 & 62.1 & 50.2 \\ 
Grounded Diffusion +  \emph{token opt.} & 
\textbf{88.3} & \textbf{75.9} & \textbf{67.1} & \textbf{95.0} & \textbf{54.3} & \textbf{89.0} & \textbf{78.1} & \textbf{85.5} & \textbf{85.6} & \textbf{79.1} & \textbf{86.6} & \textbf{73.3} \\  \hline
\end{tabular}}

\caption{Class-performance comparison of selected state-of-the-art methods employing the proposed token optimization process in OVAM.}
\label{tab:token-opt-comparison}
\vspace{-2mm}
\end{table*}

\subsection{Training a Semantic Segmentation Model}
\label{subsection:exp-training}

\textbf{Dataset Generation}. For training the semantic segmentation model, we created a large synthetic dataset by sampling 20,000 COCO captions of images containing a main object from a VOC class. Following the generation protocol from Experiment \ref{subsection:exp-1}, we generated images, and the corresponding pseudo-mask for the main class, using OVAM with optimized tokens. To automatically filter out low-quality images that could negatively impact training, we used a filter based on CLIP \cite{clip} similarity. For this filtering, we generated a CLIP embedding for each image and calculated the cosine similarity with the CLIP embedding of the text \emph{A photograph of a \textlangle classname\textrangle}. Images with lower similarity scores were more likely to contain scenes unrelated to the class and were therefore filtered out. We removed the bottom 30\% of images for each class based on these similarity scores. Additionally, we discarded images where the pseudo-mask covered more than 95\% or less than 5\% of the image area. After filtering, obtaining a final dataset with 13,484 images.

\textbf{Training setup}. Our training scheme involves Mask2former \cite{mask2former} (transformer-based) and Upernet \cite{xiao2018unified} (convolutional-based) models trained on the VOC12 dataset \cite{pascal-voc-2012}, augmented with synthetic data from OVAM, exploring two distinct scenarios. In the first, we simulate a scarcity of real data by selecting 250 images per category, totaling 5000 real images for training. In the second scenario, we utilize the entire VOC12 dataset. For experiments combining real and synthetic images, a fine-tuning protocol is applied. Specifically, starting with models trained on synthetic images, we conduct further training with an initial learning rate ten times smaller. Training settings, including a batch size of two, are adopted from MMSegmentation \cite{mmseg2020}.

\textbf{Evaluation}. Once the semantic segmentation model is trained, we follow the official VOC evaluation protocol on the VOC validation set \cite{pascal-voc-2012}.

\vspace{2mm}
\textbf{Results}. Table \ref{tab:trained_models} compiles the results of training different models with real data, synthetic data from OVAM, and their combinations. This experiment leads to two key observations: Firstly, when real data is scarce, adding synthetic data generated through OVAM can match the results obtained using double the real data. Secondly, when more real data is available, incorporating  synthetic data from OVAM can improve models performance by up to a 6.9\% in mIoU.

\begin{table*}[ht]
\centering
\setlength{\tabcolsep}{2pt} 
\small
\resizebox{\linewidth}{!}{%
\begin{tabular}{c|c|ccccccccccc|c}
\hline
\multirow{2}{*}{\begin{tabular}[c]{@{}c@{}}Model\\ (Backbone)\end{tabular}} & \multirow{2}{*}{Training set} & \multicolumn{11}{c|}{Selected classes (VOC validation set IoU \%)} & \multirow{2}{*}{mIoU \%}
\\ \cline{3-13}
&  & aeroplane & bicycle & boat & bus & car & cat & chair & cow & dog & person & tv &  \\ \hline
\multirow{4}{*}{\begin{tabular}[c]{@{}c@{}}Upernet\\ (ResNet50)\end{tabular}} & OVAM (S: 13.5K) &  45.2 & 29.5 & 19.7 & 68.6 & 46.5 & 60.5 &12.2& 55.2 & 45.0 & 50.2 &26.2& 42.1 \\
& VOC (R: 11.5K) &  75.9 & 31.3 & 33.5 & \textbf{83.6} & \textbf{76.9} & \textbf{77.6} &23.3& 49.6 & \textbf{66.8} & \textbf{73.1} & 25.9&54.3 \\
& OVAM (S: 13.5K) + VOC (R: 5K)  & 72.2 & 22.9 & 29.1 & 79.9 & 75.4& 76.8&23.3 & \textbf{58.8} & 66.2 & 72.4 & 25.3&54.4 \\
& OVAM (S: 13.5K) + VOC (R: 11.5K) & \textbf{80.8} & \textbf{47.1} & \textbf{51.8} & 52.3 & 72.5 & 71.1 & \textbf{74.5} & 21.3&57.3 & 73.0 & \textbf{39.8} & \textbf{58.1} \\\hline
\multirow{4}{*}{\begin{tabular}[c]{@{}c@{}}Mask2Former\\ (ResNet50)\end{tabular}} & OVAM (S: 13.5K) & 54.9 & 30.1 & 40.3 & 70.8 & 47.2 & 65.0 & 3.1&51.1 & 46.8 & 39.9 &0.0& 41.5 \\
 & VOC (R: 11.5K) &  88.9 & 53.0 & 62.9 & 83.0 & 83.0 & 90.9 & 27.2& \textbf{71.8} & 73.2 & 86.2 &59.6& 69.7 \\
 & OVAM (S: 13.5K) + VOC (R: 5K) &  85.4 &  53.5 &  \textbf{65.6} &   \textbf{87.7} & 82.6 & 82.6 & 26.9&66.6 & 70.1 & \textbf{86.9} & \textbf{67.0} & 69.8 \\
 & OVAM (S: 13.5K) + VOC (R: 11.5K) &  \textbf{89.7} & \textbf{63.0} & 60.6 & 84.9 & \textbf{83.4} & \textbf{91.0} & \textbf{28.5} & 69.9 & \textbf{74.2} & 86.5 & 62.8&\textbf{70.5} \\ \hline
\multirow{4}{*}{\begin{tabular}[c]{@{}c@{}}Mask2Former\\ (Swin-B)\end{tabular}}& OVAM (S: 13.5K) &  66.1 & 27.9 & 57.3 & 70.8 & 51.7 & 81.1 & 15.0& 78.9 & 68.9 & 46.4 & 0.0& 53.5 \\
&  VOC (R: 11.5K) & \textbf{94.9} & \textbf{82.0} & 81.2 & 91.5 & 90.3 & \textbf{95.4} & 42.1&85.4 & \textbf{89.4} & 90.3 & \textbf{86.7} & 82.8 \\
& OVAM (S: 13.5K) + VOC (R: 5K) &  91.7 & 79.7 &  79.1 &  \textbf{95.3} &  89.6 &  93.5 & \textbf{44.9} & 91.8 &  89.0 &  90.6 & 82.6&82.9 \\
& OVAM (S: 13.5K) + VOC (R: 11.5K) &  94.7 & 78.2 & \textbf{81.4} & 93.9 & \textbf{91.3} & 94.2 &41.7& \textbf{94.2} &88.2 & \textbf{90.6} & 79.7& \textbf{83.8} \\ \hline
\end{tabular}}
\caption{Model Performance Metrics on the VOC Validation Set\cite{pascal-voc-2012}. In the table, \emph{S} represents the number of OVAM-generated synthetic training samples, while \emph{R} denotes the count  of real training samples from the VOC dataset. Best results are indicated in bold.}
\label{tab:trained_models}
\vspace{-1mm}
\end{table*}

\subsection{Ablation Study}

To understand the impact of different components in OVAM, we conducted several ablation studies, each replicating the evaluation from Experiment \ref{subsection:exp-1} to analyze the effects of post-processing, layer, and time step selection.

\textbf{Post-processing effect}. Table \ref{tab:combinations} summarizes the findings from our study to quantify the impact of post-processing stages — specifically, the application of dCRF and self-attention refinement — on the creation of OVAM masks.

\begin{table}[ht]
\centering
\setlength{\tabcolsep}{2pt}
\small
\resizebox{\linewidth}{!}{%
\begin{tabular}{cc||cc|cc}
\hline
\multicolumn{2}{c||}{Post-processing} & \multicolumn{2}{c|}{VOC-sim (\% mIoU)} & \multicolumn{2}{c}{COCO-cap (\% mIoU)} \\ \hline
Self-Att. & dCRF & w/o opt. & w/ opt. & w/o opt. & w/ opt. \\ \hline
 & & 66.2 & 79.7 & 48.4 & 66.1 \\ 
 \cmark & & 67.7 & 80.0 & 55.0 & 66.3 \\ 
& \cmark & 68.4 & 81.9 & 50.7 & 69.0 \\ \hline
\cmark & \cmark & \textbf{70.4} & \textbf{82.5} & \textbf{58.2} & \textbf{69.2} \\ \hline
\end{tabular}}
\caption{Post-processing ablation study for the OVAM proposal.}
\label{tab:combinations}
\end{table}

\textbf{Effect of layer selection}. The denoising UNet of Stable Diffusion 1.5, use for the experiments, contain attention block of different resolution: 64x64, 32x32 and 16x16. In our proposed OVAM implementation, we aggregate attention from all resolutions. To quantify the impact of different blocks, we repeat the evaluation using different combinations of blocks. Results are compiled in Table \ref{tab:unet_blocks_mious}.
We observe that 16x16 blocks have better performance, which can be explained by the fact that they belong to a deeper part of the UNET\cite{UNET}, in charge of processing mainly semantic information.

\begin{table}[ht]
\centering
\small
\setlength{\tabcolsep}{2pt}
\begin{tabular}{p{0.7cm}p{0.7cm}p{0.7cm}||cc|cc}
\hline
\multicolumn{3}{c||}{Block resolution} & \multicolumn{2}{c|}{VOC-sim (\% mIoU)} & \multicolumn{2}{c}{COCO-cap (\% mIoU)} \\
\cline{1-7}
\centering 64 & \centering 32 & \centering\arraybackslash 16 & w/o opt. & w/ opt. & w/o opt. & w/ opt. \\
\hline
\centering\cmark &  &  &  40.2 & 43.5 & 26.2 & 24.4 \\
 & \centering\cmark &  &  30.5 & 71.7 & 16.1 & 56.0 \\
 &  & \centering\arraybackslash\cmark &  \textbf{71.5} & 78.9 & 57.0 & 68.2	 \\ \hline
\centering\cmark & \centering\cmark &  &  50.5 & 74.9 & 37.2 & 53.6 \\
\centering\cmark &  & \centering\arraybackslash\cmark &  70.5 & 82.2 & \textbf{60.3} & \textbf{70.3} \\
 & \centering\cmark & \centering\arraybackslash\cmark &  67.0 & 80.7 & 53.5 & 68.2 \\ \hline
\centering\cmark & \centering\cmark & \centering\arraybackslash\cmark &  70.4 & \textbf{82.5} & 58.2 & 69.2 \\
\hline
\end{tabular}
\caption{Cross-Attention blocks ablation study. }
\label{tab:unet_blocks_mious}
\vspace{-5mm}
\end{table}

\begin{figure}[ht]
    \centering
    \includegraphics[width=\columnwidth]{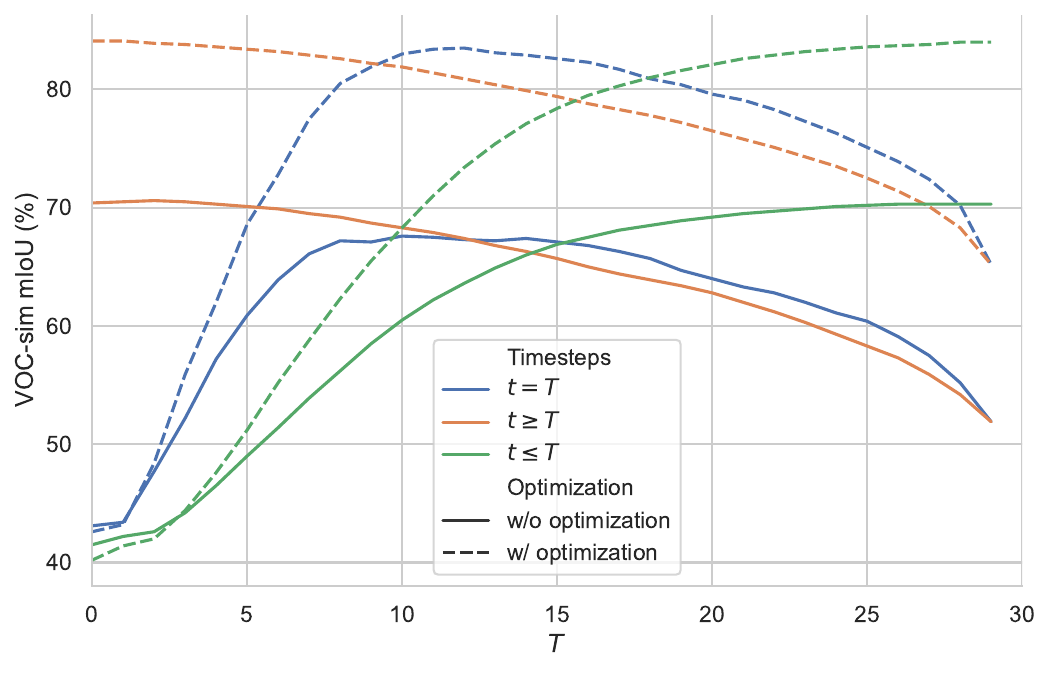}
    \vspace{-8mm}
    \caption{Time-step selection study for the OVAM proposal.}
    \label{fig:epoch-ablation}
    \vspace{-5mm}
\end{figure}

\textbf{Effect of denoising time step selection}. In our method, we aggregate attention across all time steps of the denoising process. We explored the impact of three time step selection strategies: choosing a single time step \( t=T \), selecting the initial time steps \( t \leq T \), and opting for the latter time steps \( t \geq T \). Figure \ref{fig:epoch-ablation} illustrates the results of this ablation study, varying the number of selected time steps throughout the process for these three strategies. The findings indicate that aggregating across all time steps yields the best performance. Moreover, when token optimization is used, a similar level of performance can be achieved by only extracting attentions from \( t=12 \), located at the midpoint of the diffusion process.

\section{Conclusions}
\label{sec:conclusions}

In conclusion, our work introduces Open-Vocabulary Attention Maps (OVAM), extending text-to-image diffusion models like Stable Diffusion for generating synthetic images with semantic segmentation pseudo-masks through open vocabulary descriptors. Our approach adapts existing Stable Diffusion-based segmentation methods, which were previously limited to text prompt-linked masks, to recognize any arbitrary word. Moreover, our token optimization technique notably enhances the precision of attention maps for class segmentation. Experimental results show that this optimization leads to significant performance gains, increasing OVAM pseudo-masks by +12.2 mIoU and improving other diffusion-based pseudo-mask generation methods by as much as +24.5 mIoU.

Moreover, we demonstrate the practical value of OVAM-generated synthetic data for training semantic segmentation models. When this data is used for training, models with half the amount of real data can achieve competitive results on the VOC Challenge, comparable to those trained with the full set of real data. Furthermore, when combined with the entirety of real data, certain models exhibit performance improvements of up to 6.9\% in mIoU.

Our findings affirm the viability of OVAM not only in enhancing existing diffusion-based segmentation methods but also as a valuable approach for synthetic data generation to train robust semantic segmentation models.

\vspace{5mm}
\textbf{Acknowledgement} This work has been partially supported by the SEGA-CV (TED2021-131643A-I00) and the HVD (PID2021-125051OB-I00) projects of the Ministerio de Ciencia e Innovación of the Spanish Government.

{\small
\bibliographystyle{ieee_fullname}
\bibliography{egbib}

\begin{thebibliography}{10}\itemsep=-1pt

\bibitem{breakaSCENE}
Omri Avrahami, Kfir Aberman, Ohad Fried, Daniel Cohen-Or, and Dani Lischinski.
\newblock Break-a-scene: Extracting multiple concepts from a single image.
\newblock {\em arXiv preprint arXiv:2305.16311}, 2023.

\bibitem{AttendAndExcite}
Hila Chefer, Yuval Alaluf, Yael Vinker, Lior Wolf, and Daniel Cohen-Or.
\newblock Attend-and-excite: Attention-based semantic guidance for text-to-image diffusion models.
\newblock {\em arXiv preprint arXiv:2301.13826}, 2023.

\bibitem{cocoCaptions}
Xinlei Chen, Hao Fang, Tsung-Yi Lin, Ramakrishna Vedantam, Saurabh Gupta, Piotr Doll{\'a}r, and C~Lawrence Zitnick.
\newblock Microsoft coco captions: Data collection and evaluation server.
\newblock {\em arXiv preprint arXiv:1504.00325}, 2015.

\bibitem{mask2former}
Bowen Cheng, Ishan Misra, Alexander~G. Schwing, Alexander Kirillov, and Rohit Girdhar.
\newblock Masked-attention mask transformer for universal image segmentation.
\newblock In {\em IEEE/CVF Conference on Computer Vision and Pattern Recognition (CVPR)}, pages 1280--1289, 2022.

\bibitem{BertLookAt}
Kevin Clark, Urvashi Khandelwal, Omer Levy, and Christopher~D. Manning.
\newblock What does {BERT} look at? an analysis of {BERT}{'}s attention.
\newblock In {\em BlackboxNLP Workshop on Analyzing and Interpreting Neural Networks for NLP}, pages 276--286, 2019.

\bibitem{Couairon_2023_ICCV}
Guillaume Couairon, Marl\`ene Careil, Matthieu Cord, St\'ephane Lathuili\`ere, and Jakob Verbeek.
\newblock Zero-shot spatial layout conditioning for text-to-image diffusion models.
\newblock In {\em IEEE/CVF International Conference on Computer Vision (ICCV)}, pages 2174--2183, 2023.

\bibitem{DiffusionBeatGans}
Prafulla Dhariwal and Alexander Nichol.
\newblock Diffusion models beat gans on image synthesis.
\newblock In {\em Conference on Neural Information Processing Systems (NIPS)}, volume~34, pages 8780--8794, 2021.

\bibitem{ding2021decoupling}
Jian Ding, Nan Xue, Gui-Song Xia, and Dengxin Dai.
\newblock Decoupling zero-shot semantic segmentation.
\newblock In {\em IEEE/CVF Conference on Computer Vision and Pattern Recognition (CVPR)}, pages 11583--11592, 2022.

\bibitem{pascal-voc-2012}
M. Everingham, L. Van~Gool, C.~K.~I. Williams, J. Winn, and A. Zisserman.
\newblock The {PASCAL} {V}isual {O}bject {C}lasses {C}hallenge 2012 {(VOC2012)} {R}esults.
\newblock \url{http://www.pascal-network.org/challenges/VOC/voc2012/workshop/index.html}.

\bibitem{textualInversion}
Rinon Gal, Yuval Alaluf, Yuval Atzmon, Or Patashnik, Amit~Haim Bermano, Gal Chechik, and Daniel Cohen-Or.
\newblock An image is worth one word: Personalizing text-to-image generation using textual inversion.
\newblock In {\em International Conference on Learning Representations (ICLR)}, 2023.

\bibitem{dCRFImplementation}
{Healthcare Intelligence Laboratory}.
\newblock {SimpleCRF}.
\newblock \url{https://github.com/HiLab-git/SimpleCRF}, 2017.

\bibitem{hedlin2023unsupervised}
Eric Hedlin, Gopal Sharma, Shweta Mahajan, Hossam Isack, Abhishek Kar, Andrea Tagliasacchi, and Kwang~Moo Yi.
\newblock Unsupervised semantic correspondence using stable diffusion.
\newblock In {\em Conference on Neural Information Processing Systems (NIPS)}, 2023.

\bibitem{PromptToPrompt}
Amir Hertz, Ron Mokady, Jay Tenenbaum, Kfir Aberman, Yael Pritch, and Daniel Cohen-Or.
\newblock Prompt-to-prompt image editing with cross attention control.
\newblock {\em arXiv preprint arXiv:2208.01626}, 2022.

\bibitem{DenoisingDiffusion}
Jonathan Ho, Ajay Jain, and Pieter Abbeel.
\newblock Denoising diffusion probabilistic models.
\newblock In {\em Conference on Neural Information Processing Systems (NIPS)}, volume~33, pages 6840--6851, 2020.

\bibitem{lossesSurvey}
Shruti Jadon.
\newblock A survey of loss functions for semantic segmentation.
\newblock In {\em IEEE Conference on Computational Intelligence in Bioinformatics and Computational Biology (CIBCB)}, pages 1--7, 2020.

\bibitem{ovdiff}
Laurynas Karazija, Iro Laina, Andrea Vedaldi, and Christian Rupprecht.
\newblock Diffusion models for zero-shot open-vocabulary segmentation.
\newblock {\em arXiv preprint arXiv:2306.09316}, 2023.

\bibitem{crf}
Philipp Kr\"{a}henb\"{u}hl and Vladlen Koltun.
\newblock Efficient inference in fully connected crfs with gaussian edge potentials.
\newblock In {\em Conference on Neural Information Processing Systems (NIPS)}, volume~24, pages 109--117, 2011.

\bibitem{lsegplus}
Boyi Li, Kilian~Q Weinberger, Serge Belongie, Vladlen Koltun, and Rene Ranftl.
\newblock Language-driven semantic segmentation.
\newblock In {\em International Conference on Learning Representations (ICLR)}, 2022.

\bibitem{li2023grounded}
Ziyi Li, Qinye Zhou, Xiaoyun Zhang, Ya Zhang, Yanfeng Wang, and Weidi Xie.
\newblock Open-vocabulary object segmentation with diffusion models.
\newblock In {\em IEEE/CVF International Conference on Computer Vision (ICCV)}, pages 7667--7676, 2023.

\bibitem{ovseg}
Feng Liang, Bichen Wu, Xiaoliang Dai, Kunpeng Li, Yinan Zhao, Hang Zhang, Peizhao Zhang, Peter Vajda, and Diana Marculescu.
\newblock Open-vocabulary semantic segmentation with mask-adapted {CLIP}.
\newblock In {\em IEEE/CVF Conference on Computer Vision and Pattern Recognition (CVPR)}, pages 7061--7070, 2023.

\bibitem{Lin_2019_ICCV}
Hubert Lin, Paul Upchurch, and Kavita Bala.
\newblock Block annotation: Better image annotation with sub-image decomposition.
\newblock In {\em IEEE/CVF International Conference on Computer Vision (ICCV)}, 2019.

\bibitem{instaflow}
Xingchao Liu, Xiwen Zhang, Jianzhu Ma, Jian Peng, and Qiang Liu.
\newblock Instaflow: One step is enough for high-quality diffusion-based text-to-image generation.
\newblock {\em arXiv preprint arXiv:2309.06380}, 2023.

\bibitem{luo2023dhf}
Grace Luo, Lisa Dunlap, Dong~Huk Park, Aleksander Holynski, and Trevor Darrell.
\newblock Diffusion hyperfeatures: Searching through time and space for semantic correspondence.
\newblock In {\em Conference on Neural Information Processing Systems (NIPS)}, 2023.

\bibitem{mmseg2020}
{MMSegmentation Contributors}.
\newblock {MMSegmentation}: Openmmlab semantic segmentation toolbox and benchmark, 2020.
\newblock \url{https://github.com/open-mmlab/mmsegmentation}.

\bibitem{T2IAdapter}
Chong Mou, Xintao Wang, Liangbin Xie, Jing Zhang, Zhongang Qi, Ying Shan, and Xiaohu Qie.
\newblock T2i-adapter: Learning adapters to dig out more controllable ability for text-to-image diffusion models.
\newblock {\em arXiv preprint arXiv:2302.08453}, 2023.

\bibitem{ni2023refdiff}
Minheng Ni, Yabo Zhang, Kailai Feng, Xiaoming Li, Yiwen Guo, and Wangmeng Zuo.
\newblock Ref-diff: Zero-shot referring image segmentation with generative models.
\newblock {\em arXiv preprint arXiv:2308.16777}, 2023.

\bibitem{glide}
Alexander~Quinn Nichol, Prafulla Dhariwal, Aditya Ramesh, Pranav Shyam, Pamela Mishkin, Bob Mcgrew, Ilya Sutskever, and Mark Chen.
\newblock {GLIDE}: Towards photorealistic image generation and editing with text-guided diffusion models.
\newblock In {\em International Conference on Machine Learning (ICML)}, volume 162, pages 16784--16804, 2022.

\bibitem{park2023understanding}
Yong-Hyun Park, Mingi Kwon, Jaewoong Choi, Junghyo Jo, and Youngjung Uh.
\newblock Understanding the latent space of diffusion models through the lens of riemannian geometry.
\newblock In {\em Conference on Neural Information Processing Systems (NIPS)}, 2023.

\bibitem{imageTranslation}
Gaurav Parmar, Krishna Kumar~Singh, Richard Zhang, Yijun Li, Jingwan Lu, and Jun-Yan Zhu.
\newblock Zero-shot image-to-image translation.
\newblock In {\em Special Interest Group on Computer Graphics and Interactive Techniques (SIGGRAPH)}, 2023.

\bibitem{Pytorch}
Adam Paszke, Sam Gross, Francisco Massa, Adam Lerer, James Bradbury, Gregory Chanan, Trevor Killeen, Zeming Lin, Natalia Gimelshein, Luca Antiga, Alban Desmaison, Andreas Kopf, Edward Yang, Zachary DeVito, Martin Raison, Alykhan Tejani, Sasank Chilamkurthy, Benoit Steiner, Lu Fang, Junjie Bai, and Soumith Chintala.
\newblock Pytorch: An imperative style, high-performance deep learning library.
\newblock In {\em Conference on Neural Information Processing Systems (NIPS)}, pages 8024--8035, 2019.

\bibitem{wuerstchen}
Pablo Pernias, Dominic Rampas, and Marc Aubreville.
\newblock Wuerstchen: Efficient pretraining of text-to-image models.
\newblock {\em arXiv preprint arXiv:2306.00637}, 2023.

\bibitem{PNVR_2023_ICCV}
Koutilya PNVR, Bharat Singh, Pallabi Ghosh, Behjat Siddiquie, and David Jacobs.
\newblock Ld-znet: A latent diffusion approach for text-based image segmentation.
\newblock In {\em IEEE/CVF International Conference on Computer Vision (ICCV)}, pages 4157--4168, 2023.

\bibitem{sdxl}
Dustin Podell, Zion English, Kyle Lacey, Andreas Blattmann, Tim Dockhorn, Jonas Müller, Joe Penna, and Robin Rombach.
\newblock Sdxl: Improving latent diffusion models for high-resolution image synthesis.
\newblock {\em arXiv preprint arXiv:2307.01952}, 2023.

\bibitem{freeseg}
Jie Qin, Jie Wu, Pengxiang Yan, Ming Li, Ren Yuxi, Xuefeng Xiao, Yitong Wang, Rui Wang, Shilei Wen, Xin Pan, et~al.
\newblock Freeseg: Unified, universal and open-vocabulary image segmentation.
\newblock {\em IEEE/CVF Conference on Computer Vision and Pattern Recognition (CVPR)}, pages 19446--19455, 2023.

\bibitem{DatasetDiffusion}
Nguyen Quang~Ho, Vu Truong~Tuan, Tran Anh~Tuan, and Nguyen Khoi.
\newblock Dataset diffusion: Diffusion-based synthetic data generation for pixel-level semantic segmentation.
\newblock In {\em Conference on Neural Information Processing Systems (NIPS)}, 2023.

\bibitem{clip}
Alec Radford, Jong~Wook Kim, Chris Hallacy, A. Ramesh, Gabriel Goh, Sandhini Agarwal, Girish Sastry, Amanda Askell, Pamela Mishkin, Jack Clark, Gretchen Krueger, and Ilya Sutskever.
\newblock Learning transferable visual models from natural language supervision.
\newblock In {\em International Conference on Machine Learning (ICML)}, 2021.

\bibitem{t5}
Colin Raffel, Noam Shazeer, Adam Roberts, Katherine Lee, Sharan Narang, Michael Matena, Yanqi Zhou, Wei Li, and Peter~J. Liu.
\newblock Exploring the limits of transfer learning with a unified text-to-text transformer.
\newblock {\em Journal of Machine Learning Research (JMLR)}, 21(140):1--67, 2020.

\bibitem{Dalle2}
Aditya Ramesh, Prafulla Dhariwal, Alex Nichol, Casey Chu, and Mark Chen.
\newblock Hierarchical text-conditional image generation with clip latents.
\newblock {\em arXiv preprint arXiv:2204.06125}, 2022.

\bibitem{kandinsky}
Anton Razzhigaev, Arseniy Shakhmatov, Anastasia Maltseva, Vladimir Arkhipkin, Igor Pavlov, Ilya Ryabov, Angelina Kuts, Alexander Panchenko, Andrey Kuznetsov, and Denis Dimitrov.
\newblock Kandinsky: an improved text-to-image synthesis with image prior and latent diffusion.
\newblock {\em arXiv preprint arXiv:2310.03502}, 2023.

\bibitem{sd}
Robin Rombach, Andreas Blattmann, Dominik Lorenz, Patrick Esser, and Bj\"orn Ommer.
\newblock High-resolution image synthesis with latent diffusion models.
\newblock In {\em IEEE/CVF Conference on Computer Vision and Pattern Recognition (CVPR)}, pages 10684--10695, 2022.

\bibitem{UNET}
Olaf Ronneberger, Philipp Fischer, and Thomas Brox.
\newblock U-net: Convolutional networks for biomedical image segmentation.
\newblock In {\em International Conference on Medical Image Computing and ComputerAssisted Intervention (MICCAI)}, pages 234--241, 2015.

\bibitem{IMAGEN}
Chitwan Saharia, William Chan, Saurabh Saxena, Lala Li, Jay Whang, Emily~L Denton, Kamyar Ghasemipour, Raphael Gontijo~Lopes, Burcu Karagol~Ayan, Tim Salimans, Jonathan Ho, David~J Fleet, and Mohammad Norouzi.
\newblock Photorealistic text-to-image diffusion models with deep language understanding.
\newblock In {\em Conference on Neural Information Processing Systems (NIPS)}, volume~35, pages 36479--36494, 2022.

\bibitem{Schwartz_2023_ICCV}
Idan Schwartz, V\'esteinn Sn{\ae}bjarnarson, Hila Chefer, Serge Belongie, Lior Wolf, and Sagie Benaim.
\newblock Discriminative class tokens for text-to-image diffusion models.
\newblock In {\em IEEE/CVF International Conference on Computer Vision (ICCV)}, pages 22725--22735, 2023.

\bibitem{NonequilibriumThermodynamics}
Jascha Sohl-Dickstein, Eric Weiss, Niru Maheswaranathan, and Surya Ganguli.
\newblock Deep unsupervised learning using nonequilibrium thermodynamics.
\newblock In {\em International Conference on Machine Learning (ICML)}, volume~37, pages 2256--2265, 2015.

\bibitem{tang2023dift}
Luming Tang, Menglin Jia, Qianqian Wang, Cheng~Perng Phoo, and Bharath Hariharan.
\newblock Emergent correspondence from image diffusion.
\newblock {\em arXiv preprint arXiv:2306.03881}, 2023.

\bibitem{DAAM}
Raphael Tang, Linqing Liu, Akshat Pandey, Zhiying Jiang, Gefei Yang, Karun Kumar, Pontus Stenetorp, Jimmy Lin, and Ferhan Ture.
\newblock What the {DAAM}: Interpreting stable diffusion using cross attention.
\newblock In {\em Annual Meeting of the Association for Computational Linguistics (ACL)}, volume~1, pages 5644--5659, 2023.

\bibitem{KeyLockedRank}
Yoad Tewel, Rinon Gal, Gal Chechik, and Yuval Atzmon.
\newblock Key-locked rank one editing for text-to-image personalization.
\newblock {\em Special Interest Group on Computer Graphics and Interactive Techniques (SIGGRAPH)}, 2023.

\bibitem{diffuseAttendSegment}
Junjiao Tian, Lavisha Aggarwal, Andrea Colaco, Zsolt Kira, and Mar Gonzalez-Franco.
\newblock Diffuse, attend, and segment: Unsupervised zero-shot segmentation using stable diffusion.
\newblock {\em arXiv preprint arXiv:2308.12469}, 2023.

\bibitem{Tumanyan_2023_CVPR}
Narek Tumanyan, Michal Geyer, Shai Bagon, and Tali Dekel.
\newblock Plug-and-play diffusion features for text-driven image-to-image translation.
\newblock In {\em IEEE/CVF Conference on Computer Vision and Pattern Recognition (CVPR)}, pages 1921--1930, 2023.

\bibitem{AttentionIsAllYouNeed}
Ashish Vaswani, Noam Shazeer, Niki Parmar, Jakob Uszkoreit, Llion Jones, Aidan~N Gomez, {\L}ukasz Kaiser, and Illia Polosukhin.
\newblock Attention is all you need.
\newblock In {\em Conference on Neural Information Processing Systems (NIPS)}, volume~30, pages 6000--6010, 2017.

\bibitem{wang2023ovvg}
Chunlei Wang, Wenquan Feng, Xiangtai Li, Guangliang Cheng, Shuchang Lyu, Binghao Liu, Lijiang Chen, and Qi Zhao.
\newblock Ov-vg: A benchmark for open-vocabulary visual grounding.
\newblock {\em arXiv preprint arXiv:2310.14374}, 2023.

\bibitem{SecretSegmenter}
Jinglong Wang, Xiawei Li, Jing Zhang, Qingyuan Xu, Qin Zhou, Qian Yu, Lu Sheng, and Dong Xu.
\newblock Diffusion model is secretly a training-free open vocabulary semantic segmenter.
\newblock {\em arXiv preprint arXiv:2309.02773}, 2023.

\bibitem{hardprompts}
Yuxin Wen, Neel Jain, John Kirchenbauer, Micah Goldblum, Jonas Geiping, and Tom Goldstein.
\newblock Hard prompts made easy: Gradient-based discrete optimization for prompt tuning and discovery.
\newblock In {\em Conference on Neural Information Processing Systems (NIPS)}, 2023.

\bibitem{AttentionNotNot}
Sarah Wiegreffe and Yuval Pinter.
\newblock Attention is not not explanation.
\newblock In {\em Conference on Empirical Methods in Natural Language Processing and International Joint Conference on Natural Language Processing (EMNLP-IJCNLP)}, pages 11--20, 2019.

\bibitem{DatasetDM}
Weijia Wu, Yuzhong Zhao, Hao Chen, Yuchao Gu, Rui Zhao, Yefei He, Hong Zhou, Mike~Zheng Shou, and Chunhua Shen.
\newblock Datasetdm: Synthesizing data with perception annotations using diffusion models.
\newblock {\em Conference on Neural Information Processing Systems (NIPS)}, 2023.

\bibitem{diffumask}
Weijia Wu, Yuzhong Zhao, Mike~Zheng Shou, Hong Zhou, and Chunhua Shen.
\newblock Diffumask: Synthesizing images with pixel-level annotations for semantic segmentation using diffusion models.
\newblock {\em IEEE/CVF International Conference on Computer Vision (ICCV)}, pages 1206--1217, 2023.

\bibitem{xiao2023text}
Changming Xiao, Qi Yang, Feng Zhou, and Changshui Zhang.
\newblock From text to mask: Localizing entities using the attention of text-to-image diffusion models.
\newblock {\em arXiv preprint arXiv:2309.01369}, 2023.

\bibitem{xiao2018unified}
Tete Xiao, Yingcheng Liu, Bolei Zhou, Yuning Jiang, and Jian Sun.
\newblock Unified perceptual parsing for scene understanding.
\newblock In {\em European Conference on Computer Vision (ECCV)}, pages 418--434, 2018.

\bibitem{ODISE}
Jiarui Xu, Sifei Liu, Arash Vahdat, Wonmin Byeon, Xiaolong Wang, and Shalini De~Mello.
\newblock Open-vocabulary panoptic segmentation with text-to-image diffusion models.
\newblock In {\em IEEE/CVF Conference on Computer Vision and Pattern Recognition (CVPR)}, pages 2955--2966, 2023.

\bibitem{xu2021}
Mengde Xu, Zheng Zhang, Fangyun Wei, Yutong Lin, Yue Cao, Han Hu, and Xiang Bai.
\newblock A simple baseline for open vocabulary semantic segmentation with pre-trained vision-language model.
\newblock In {\em European Conference on Computer Vision (ECCV)}, 2021.

\bibitem{Attn2mask}
Ryota Yoshihashi, Yuya Otsuka, Kenji Doi, and Tomohiro Tanaka.
\newblock Attention as annotation: Generating images and pseudo-masks for weakly supervised semantic segmentation with diffusion.
\newblock {\em arXiv preprint arXiv:2309.01369}, 2023.

\bibitem{diffusionSurvey}
Chenshuang Zhang, Chaoning Zhang, Mengchun Zhang, and In~So Kweon.
\newblock Text-to-image diffusion models in generative {AI}: A survey.
\newblock {\em arXiv preprint arXiv:2303.07909}, 2023.

\bibitem{zhang2023tale}
Junyi Zhang, Charles Herrmann, Junhwa Hur, Luisa~Polania Cabrera, Varun Jampani, Deqing Sun, and Ming-Hsuan Yang.
\newblock A tale of two features: Stable diffusion complements dino for zero-shot semantic correspondence.
\newblock {\em arXiv preprint arXiv:2305.15347}, 2023.

\bibitem{zhang2023realworld}
Yuechen ZHANG, Jinbo Xing, Eric Lo, and Jiaya Jia.
\newblock Real-world image variation by aligning diffusion inversion chain.
\newblock In {\em Conference on Neural Information Processing Systems (NIPS)}, 2023.

\end{thebibliography}
}

\clearpage

\setcounter{figure}{0}
\let\oldthefigure\thefigure
\renewcommand{\thefigure}{S\oldthefigure}
\setcounter{table}{0}
\let\oldthetable\thetable
\renewcommand{\thetable}{S\oldthetable}

\onecolumn
\begin{center}
\large{\bf{Supplementary}}    
\end{center}
\appendix


This supplementary material offers additional details and extended information on the token optimization used in the experiments of the paper Open-Vocabulary Attention Maps with Token Optimization for Semantic Segmentation in Diffusion Models (see Section 4), additional evaluation, and qualitative examples of the results.


\section{Implementation Details}

\subsection{Token Optimization via OVAM}

\textbf{Synthetic Images and Ground Truth Generation}.
To optimize a token for each of the 20 classes of the VOC Challenge \cite{pascal-voc-2012}, we generated one image per class using the text prompt \emph{A photograph of a \textlangle classname\textrangle}. We employed Stable Diffusion 1.5\footnote{Stable Diffusion 1.5 model card: \url{https://huggingface.co/runwayml/stable-diffusion-v1-5} (Accessed November 2023)}, the same architecture used for other OVAM experiments. We utilized 30 time steps for image generation, and default parameters of the model. The target class object in the generated image was manually annotated at a resolution of 512x512.

\textbf{Initializing Token Optimization}. 
The optimization procedure, along with other components of OVAM, is implemented in PyTorch \cite{Pytorch}. We use gradient descent to optimize tokens. Initially, the Stable Diffusion 1.5 Text Encoder (CLIP ViT-L/14 \cite{clip}) is employed to encode the text prompt \emph{A photograph of a \textlangle  classname\textrangle}. This encoder produces tokens with shape 1x768 and includes two special characters to mark the start and end of the text: \textlangle SoT\textrangle \,and \textlangle  EoT\textrangle . We initialize an attribution prompt, \(X'\), for optimization with tokens corresponding to \textlangle SoT\textrangle  \,and the classname, forming an array of size 2x768. The \textlangle SoT\textrangle \,token is recognized for attracting background attention \cite{DAAM}.

\begin{figure}[h]
    \centering
    \begin{subfigure}[b]{0.49\textwidth}
        \centering
        \includegraphics[width=\textwidth]{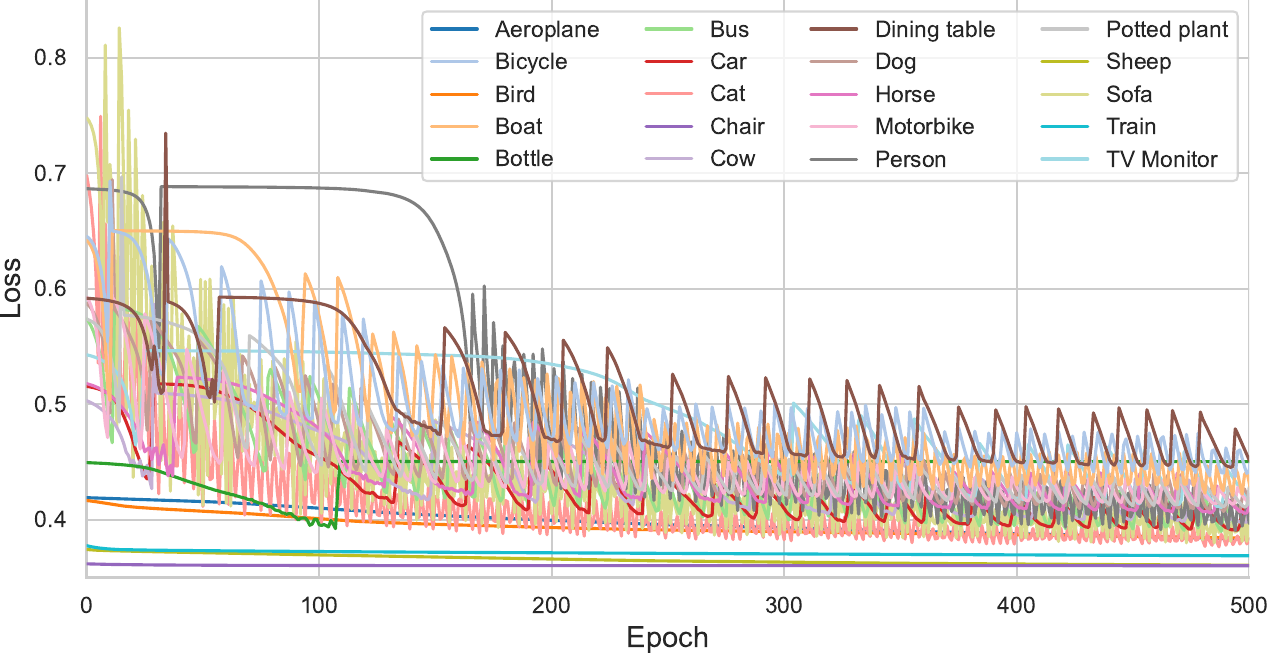}
        \caption{Losses during optimization}
        \label{fig:losses}
    \end{subfigure}
    \hfill 
    \begin{subfigure}[b]{0.49\textwidth}
        \centering
        \includegraphics[width=\textwidth]{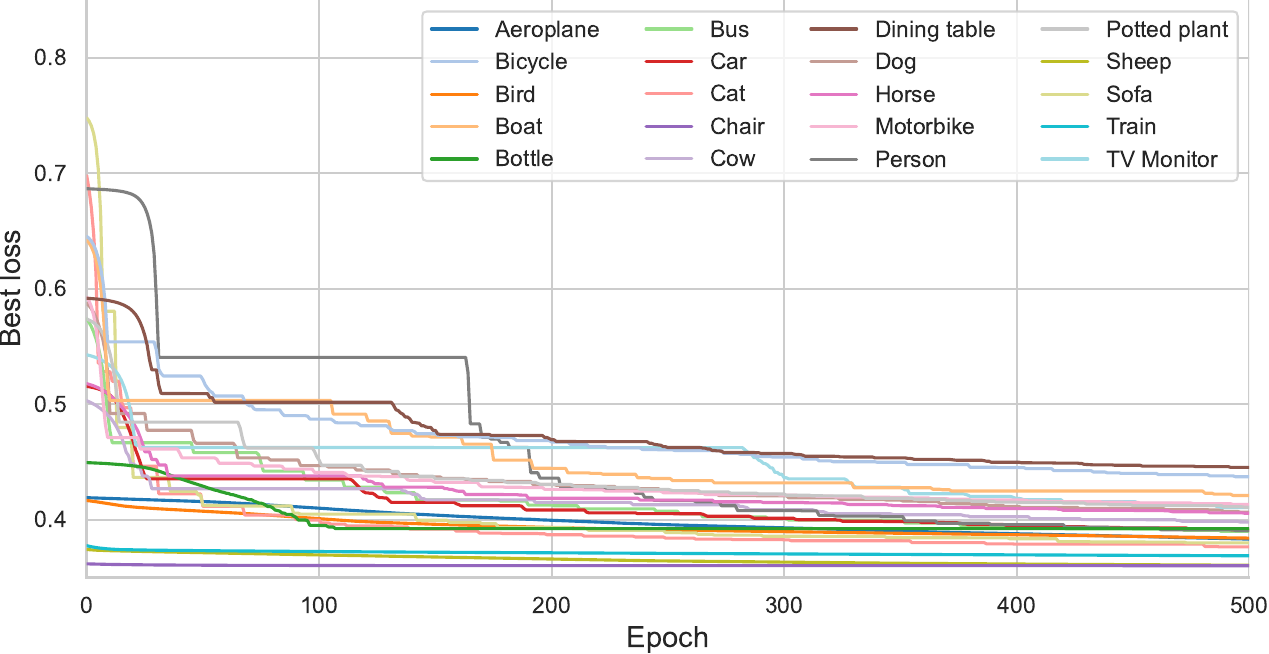}
        \caption{Best observed losses}
        \label{fig:best_losses}
    \end{subfigure}
    \caption{Losses during optimization. (a) shows the losses during the training process, and (b) presents the best losses achieved.}
    \label{fig:combined_losses}
\end{figure}

\textbf{Performing Token Optimization}.
During optimization, \(X'\) is used to generate OVAM according to the methodology outlined in the paper, resulting in two attention maps of size 2x64x64. These are scaled to an image resolution of 2x512x512 using bilinear interpolation. For each channel associated with a token, binary cross-entropy is utilized to measure the discrepancy with the annotated ground truth. The loss is then backpropagated to update \(X'\). An initial learning rate of \(\alpha=100\) is set, with a decay rate of \(\gamma=0.7\) applied every 120 steps. We run the optimization for 500 epochs, which takes less than a minute on an A40 GPU, and the best embedding is saved (Fig. \ref{fig:best_losses}). Figure \ref{fig:combined_losses} displays the learning curves for this optimization. Despite the spiked profile of the curves by class (Fig. \ref{fig:losses}), the procedure converges to values that generate accurate attention maps for all classes (see Fig. \ref{fig:s-example-optimized}).

\textbf{Use of Optimized Tokens}. The 20 tokens, each optimized using one annotated training image, are subsequently employed to generate attention maps for different images, thereby without the need for repeated optimization. The annex Sections \ref{sec:A2} and \ref{sec:A3} details the process of creating attention maps using these tokens and Section \ref{sec:B} provides qualitative examples of OVAM-generated maps. It also discusses the results of utilizing these tokens in conjunction with other methods.

\subsection{Evaluation of OVAM with Optimized Tokens}
\label{sec:A2}
\begin{figure}[ht]
    \centering
    \begin{subfigure}[b]{0.49\textwidth}
        \centering
        \includegraphics[width=\textwidth]{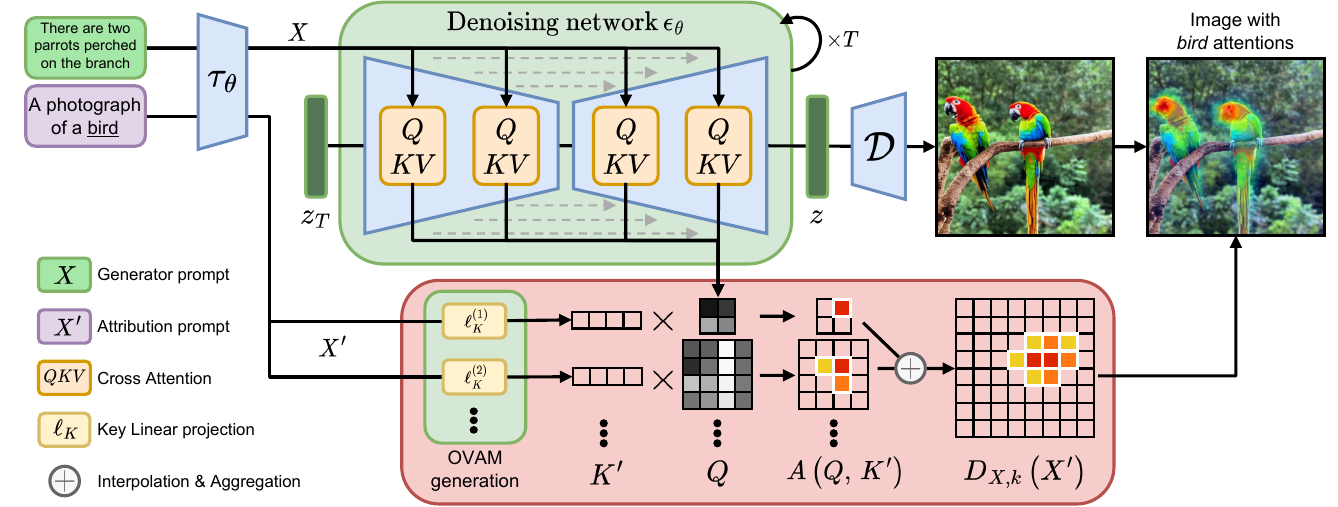}
        \caption{Evaluation of non-optimized attribution prompt}
        \label{fig:non-optimized-workflow}
    \end{subfigure}
    \hfill 
    \begin{subfigure}[b]{0.49\textwidth}
        \centering
        \includegraphics[width=\textwidth]{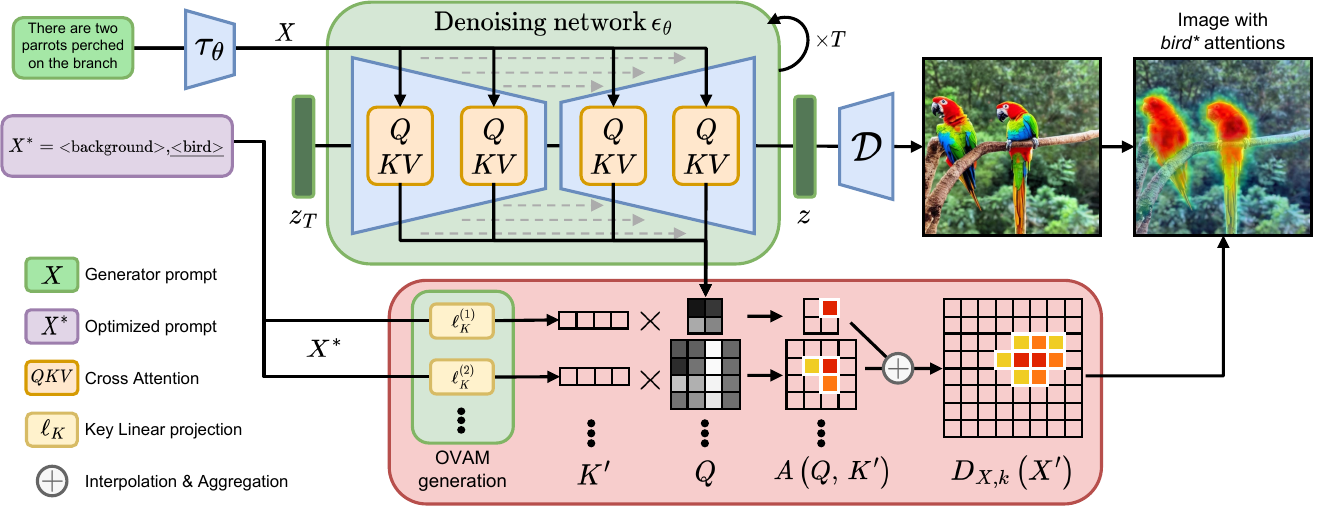}
        \caption{Evaluation of optimized prompt}
        \label{fig:optimized-workflow}
    \end{subfigure}
    \caption{Comparison between the evaluation of OVAM attention maps based on (a) a natural text description, where the attentions for the word \emph{bird} are extracted, and (b) the evaluation using an optimized token for the class \emph{bird}.}
    \label{fig:workflow-comparison}
\end{figure}

\textbf{Evaluation with Natural Text}. In the evaluation of OVAM attention maps with natural text (Fig. \ref{fig:non-optimized-workflow}), an attribution text is transformed using the Stable Diffusion 1.5 text encoder to produce a text embedding with dimensions $768 \times l_{X'}$. This embedding is then used to compute OVAM attention maps of dimensions $l_{X'} \times 64 \times 64$. Relevant maps (e.g., those corresponding to class name nouns) are extracted and resized to an image resolution of $512 \times 512$.

\textbf{Evaluation with Optimized Tokens}. For the evaluation using optimized tokens  (Fig. \ref{fig:optimized-workflow}), the input, shaped $2 \times 768$ (representing one token for the background and another for the class object), is utilized to compute two attention maps of dimensions $2 \times 64 \times 64$. The channel corresponding to the class object is selected and resized to form a $512 \times 512$ heatmap.

\begin{figure}[h]
    \centering
    \includegraphics[width=\textwidth]{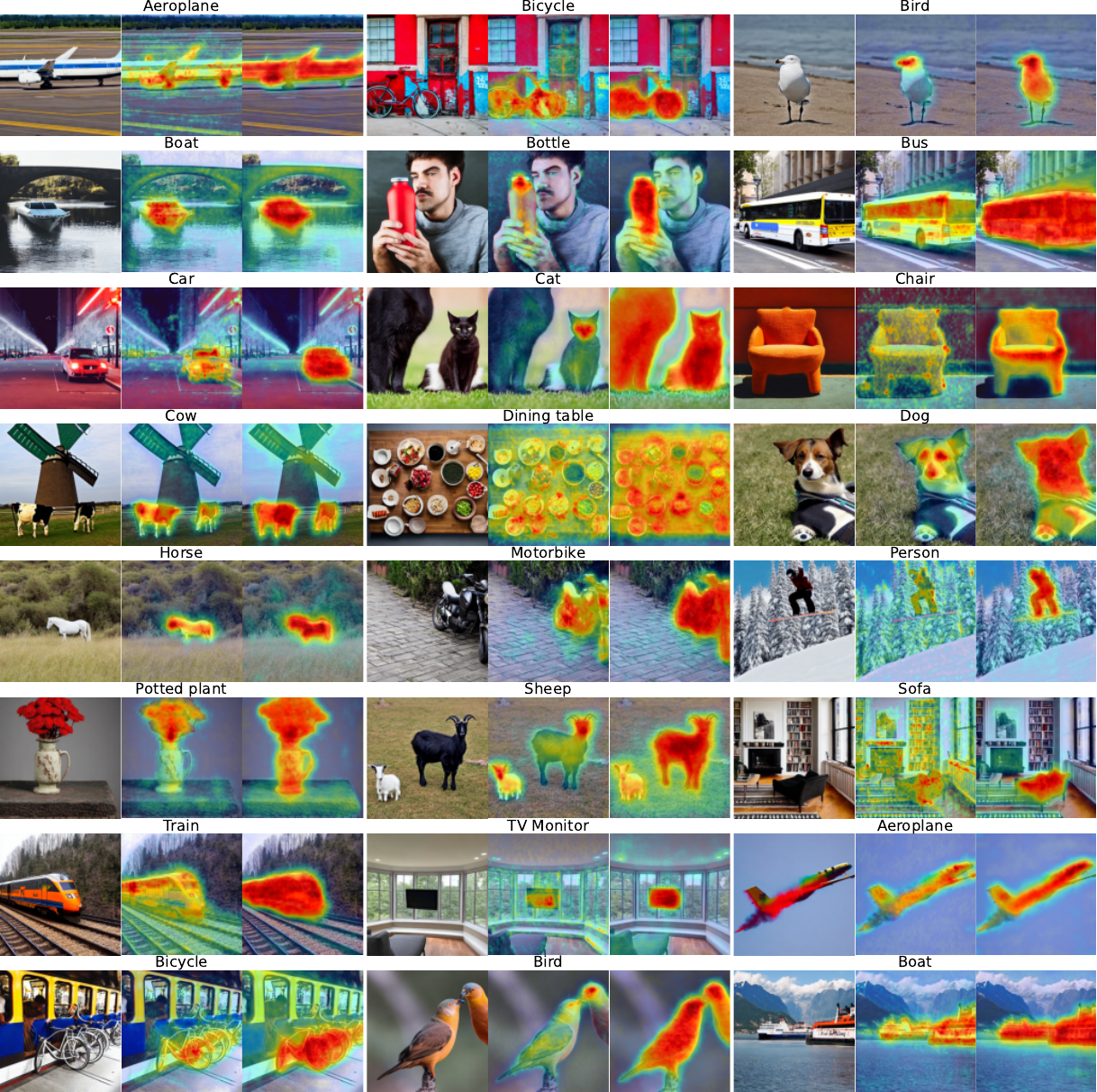}
    \caption{Qualitative Examples of synthetic images generated with Stable Diffusion 1.5 \cite{sd} and OVAM Attention Maps before binarization. For each class name, we show the obtained synthetic image (left),  the attention map generated using the class name (center) and class-specific optimized tokens (right) for each of the 20 classes from the VOC challenge \cite{pascal-voc-2012}. Images have been generated using text prompts extracted from COCO captions \cite{cocoCaptions}.}
    \label{fig:s-example-optimized}
\end{figure}

\textbf{Threshold Difference}. For binarizing maps generated from non-optimized tokens, a threshold of $\tau=0.4$ is applied, followed by self-attention post-processing and dCRF. This threshold choice is based on values used in DAAM \cite{DAAM}, a work in which OVAM's theoretical foundation is based. However, when using optimized tokens, we observe a shift in attention scale, with higher values near foreground objects (as illustrated in Fig. \ref{fig:s-example-optimized}). Preliminary experiments suggest $\tau=0.8$ as a more suitable threshold for evaluating optimized tokens.

\subsection{Evaluation of other Stable Diffusion-based works}
\label{sec:A3}
In this subsection we include further details of the evaluation of other Stable Diffusion-based works used in the experiments. Specifically, we employ DAAM \cite{DAAM}, Attn2Mask \cite{Attn2mask}, DatasetDM \cite{DatasetDM}, and Grounded Diffusion \cite{li2023grounded}.

\textbf{Grounded Diffusion Implementation Details}. Grounded Diffusion \cite{li2023grounded} extends Stable Diffusion for generating segmentation masks based on textual descriptions, by incorporating an additional trainable grounding module. This module, requiring annotated data for training, processes attentions generated during image synthesis alongside a word or token embedding. For our experiments, we utilized the official implementation available at \url{https://github.com/Lipurple/Grounded-Diffusion}, employing the weights trained with VOC classes and default setup provided by the authors. To evaluate with an optimized token, we adapted their evaluation script, allowing for direct token input instead of using a text word that is later converted into a token.

\textbf{DatasetDM Implementation Details}. DatasetDM \cite{DatasetDM} extends Stable Diffusion for various perception tasks, such as semantic segmentation, pose detection, and depth estimation. It includes a decoder that processes diffusion attentions and convolutional features. This decoder is trained using supervised examples. In our experiments, we employed DatasetDM's configuration for semantic segmentation along with the weights provided by the authors, trained for segmenting VOC classes. Official implementation used is available at \url{https://github.com/showlab/DatasetDM}. To evaluate optimized tokens, we modified their evaluation script to allow direct token input, instead of using a text word that is later converted.

\textbf{DAAM Implementation Details}. DAAM \cite{DAAM} is based on the direct extraction of cross-attentions during the synthesis process in Stable Diffusion. These attentions, extracted from all generation timesteps, blocks, and heads, are then aggregated and thresholded. We utilized the implementation available at \url{https://github.com/castorini/daam}, applying a threshold of \(\tau=0.4\), as recommended by the authors. For our experiments, we employed Stable Diffusion 1.5 with a 30-step generation process, aligning with the OVAM configuration. To evaluate DAAM in scenarios where the target class is not explicitly mentioned in the text prompt or using optimized tokens, we adapted DAAM to use OVAM attentions (similar adaptation illustrated in \ref{fig:workflow-comparison}), which provides the same result when the word is mentioned but allowing the evaluation in all cases. For optimized tokens we use a threshold $\tau=0.8$.

\textbf{Attn2Mask}. The concurrent work Attn2Mask \cite{Attn2mask} does not have any public implementation available at the time of writing this paper. Due to its similarity to OVAM without token optimization, we implemented Attn2Mask as described by the authors. For the implementation, we use Stable Diffusion 1.5 for image generation with 100 time steps. We extract cross-attentions at $t=50$ and aggregate them. The aggregated attentions are binarized with a threshold of $\tau=0.5$ and a dCRF \cite{crf} post-processing is applied using the SimpleCRF \cite{dCRFImplementation} implementation with default parameters. To evaluate optimized tokens or classes in images where the class name is not mentioned, we modify the use of cross-attention with open-vocabulary attention maps (similar adaptation illustrated in \ref{fig:workflow-comparison}). For optimized tokens, we use a threshold of $\tau=0.8$.

\clearpage
\section{Additional Experiments}

\subsection{Synthetic Data Training}
\label{subsect:additional-synthetic-training}

\begin{figure}[h]
    \centering
    \includegraphics[width=0.87\textwidth]{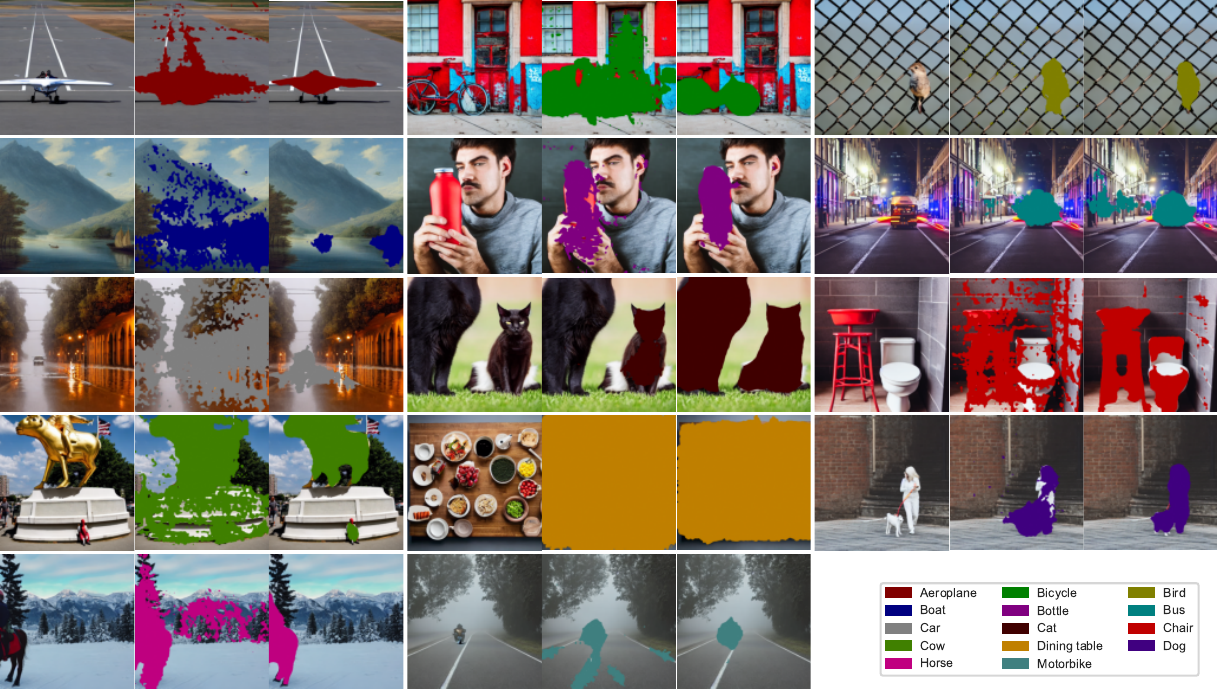}
    \caption{Qualitative Examples of \textbf{DAAM-Generated Pseudo-Masks}: Each set in the figure presents a synthetic image generated with Stable Diffusion \cite{sd} using a COCO caption \cite{cocoCaptions} (left), accompanied by a mask generated through DAAM \cite{DAAM} using VOC class names \cite{pascal-voc-2012} (center), and a mask generated using an OVAM-optimized token specific to the class (right).}
    \label{fig:daam-examples}
\end{figure}

\begin{figure}[h]
    \centering
    \includegraphics[width=0.87\textwidth]{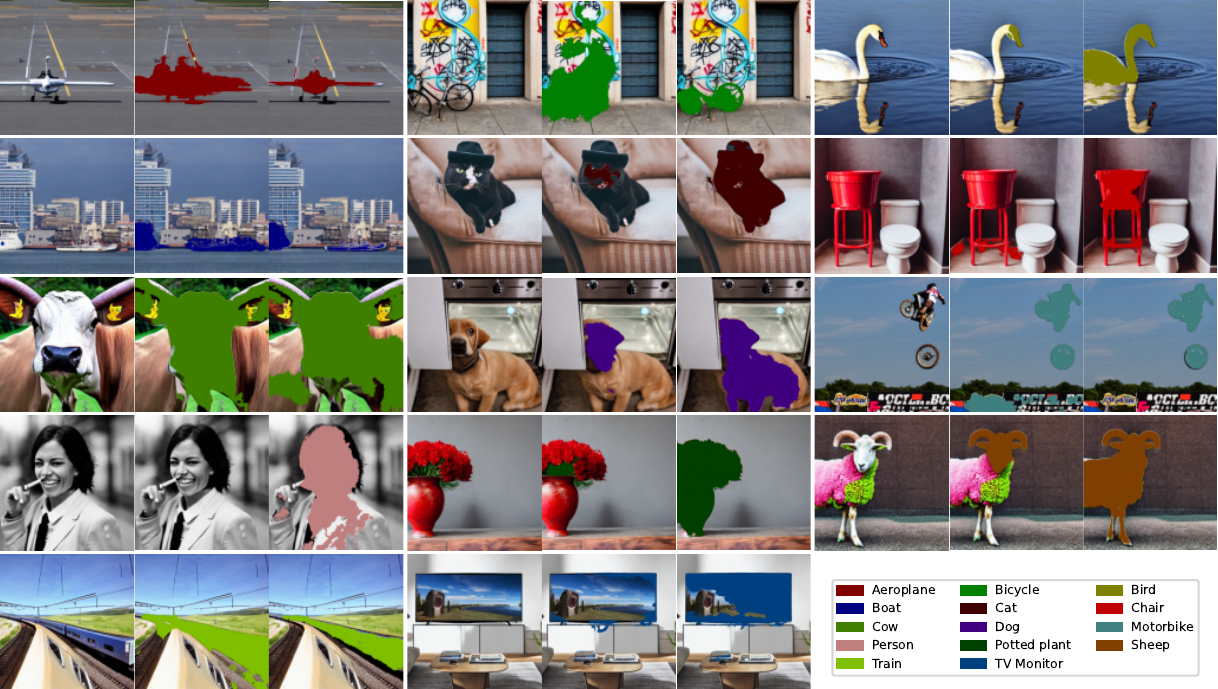}
    \caption{Qualitative Examples of \textbf{Attn2Mask-Generated Pseudo-Masks}: Each set in the figure presents a synthetic image generated with Stable Diffusion \cite{sd} using a COCO caption \cite{cocoCaptions} (left), accompanied by a mask generated through Attn2Mask \cite{Attn2mask} using VOC class names \cite{pascal-voc-2012} (center), and a mask generated using an OVAM-optimized token specific to the class (right).}
    \label{fig:attn2mask-examples}
\end{figure}

\begin{figure}[h]
    \centering
    \includegraphics[width=0.87\textwidth]{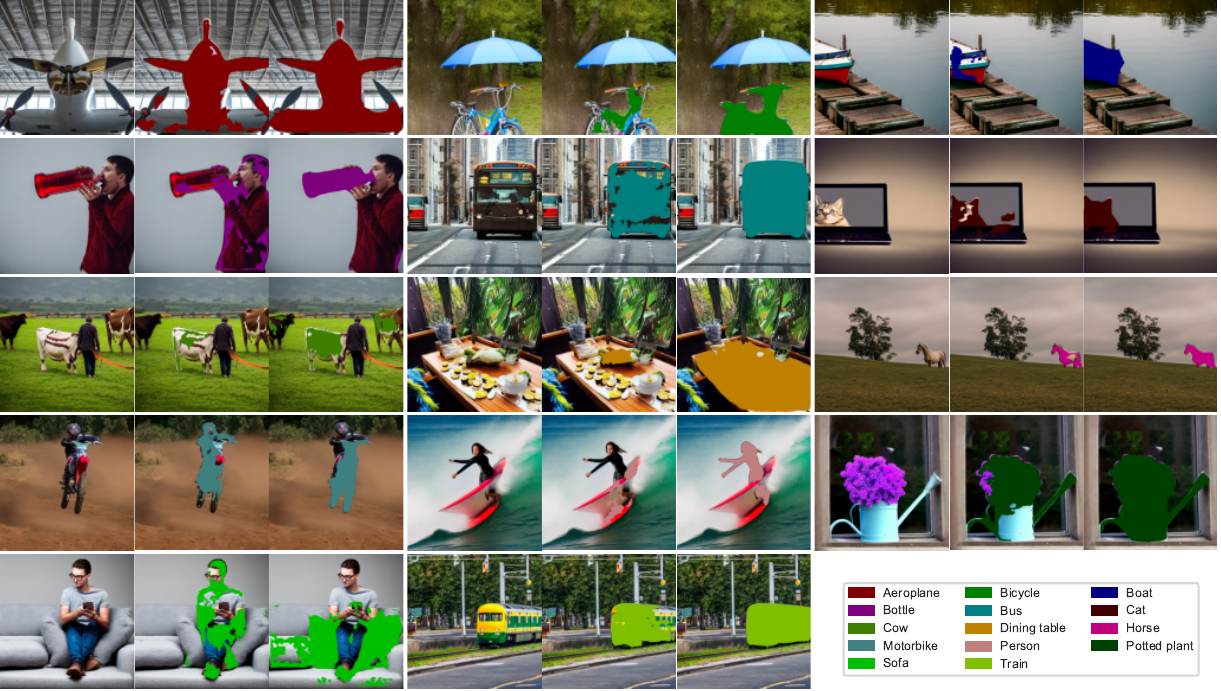}
    \caption{Qualitative Examples of \textbf{Grounded Diffusion-Generated Pseudo-Masks}: Each set in the figure presents a synthetic image generated with Stable Diffusion \cite{sd} using a COCO caption \cite{cocoCaptions} (left), accompanied by a mask generated through Grounded Diffusion \cite{li2023grounded} using VOC class names \cite{pascal-voc-2012} (center), and a mask generated using an OVAM-optimized token specific to the class (right).}
    \label{fig:grounded-diffusion-masks}
\end{figure}

\begin{figure}[h]
    \centering
    \includegraphics[width=0.87\textwidth]{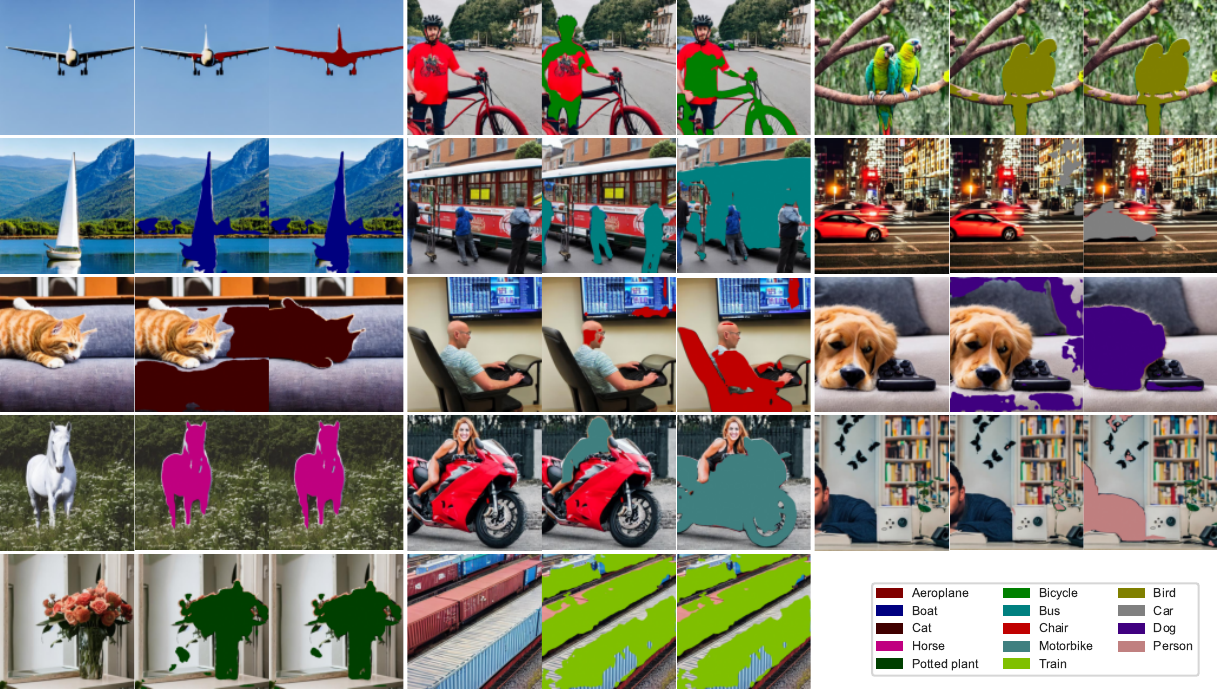}
    \caption{Qualitative Examples of \textbf{DatasetDM-Generated Pseudo-Masks}: Each set in the figure presents a synthetic image generated with Stable Diffusion \cite{sd} using a COCO caption \cite{cocoCaptions} (left), accompanied by a mask generated through DatasetDM \cite{DatasetDM} using VOC class names \cite{pascal-voc-2012} (center), and a mask generated using an OVAM-optimized token specific to the class (right). Notably, masks with non-optimized tokens sometimes segment a foreground object that does not match the intended descriptor (e.g., \emph{cat}, \emph{bus}, \emph{motorbike}). The use of optimized tokens helps in aligning DatasetDM masks more accurately with the specified objects}
    \label{fig:datasetdm-masks}
\end{figure}

Extending the evaluation of the experiment in which various semantic segmentation architectures were trained using a synthetic dataset generated by OVAM (Section 4.3), this additional experiment compares the performance of optimized tokens for generating synthetic data for semantic segmentation on the VOC Challenge \cite{pascal-voc-2012}.
To generate each dataset, 1,000 synthetic images were produced using COCO captions as prompts through various Stable Diffusion extensions: DAAM \cite{DAAM}, DatasetDM \cite{DatasetDM}, Grounded Diffusion \cite{li2023grounded}, and OVAM, to extract pseudo-masks. Subsequently, a Uppernet architecture with a ResNet-50 backbone was trained on these datasets, evaluated against the official VOC challenge protocol. This study further investigates the utility of optimized tokens: for each dataset, we extracted pseudo-masks using class names as descriptors for the VOC's 20 classes, comparing the outcomes with and without the use of optimized tokens. The incorporation of optimized tokens significantly enhanced mask quality (as evidenced in Figures \ref{fig:daam-examples} - \ref{fig:datasetdm-masks}), which, in turn, improved the performance of the trained segmentor across all classes when compared to the non-optimized approach (refer to Table \ref{tab:additional-synthetic-training-comparison}). These findings affirm the value of optimized tokens in boosting the precision of synthetically generated data by these methods, enabling effective method adaptation without additional computational costs.

\begin{table}[htpb]
\centering
\setlength{\tabcolsep}{3pt} 
\begin{tabular}{c|c|ccccccccccc|c}
\hline
\multirow{2}{*}{Method} & \multirow{2}{*}{\shortstack{Token\\Optim.}} & \multicolumn{11}{c|}{Selected classes (VOC validation set IoU \%)} & \multirow{2}{*}{mIoU} \\
\cline{3-13}
       &       & aeroplane & bicycle & bird & boat & bus & car & cat & dog & horse & person & train &  \\
\hline
\multirow{2}{*}{DAAM \cite{DAAM}} & \xmark & 27.2 & 16.6 & 41.1 & 21.9 & 56.0 & 37.3 & 32.7 & 30.6 & 25.9 & 23.1 & 38.4 & 27.5 \\
                      & \cmark & \textbf{37.0} & \textbf{28.7} & \textbf{48.8} & \textbf{28.4} & \textbf{53.5} & \textbf{54.8} & \textbf{33.4} & \textbf{31.6} & \textbf{40.9} & \textbf{24.3} & \textbf{50.3} & \textbf{32.5} \\
\hline
\multirow{2}{*}{DatasetDM\cite{DatasetDM}} & \xmark & 60.0 & 23.0 & 44.9 & 41.4 & 45.0 & 52.8 & 33.9 & 27.5 & \textbf{48.9} & 14.3 & 41.5 & 34.0 \\
                           & \cmark & \textbf{60.7} & \textbf{31.8} & \textbf{51.6} & \textbf{41.7} & \textbf{58.7} & \textbf{56.8} & \textbf{37.3} & \textbf{32.7} & \textbf{48.9} & \textbf{20.1} & \textbf{56.3} & \textbf{35.5} \\
\hline
Grounded & \xmark & 63.4 & 10.2 & 34.3 & 12.6 & 17.2 & 20.6 & 38.2 & 38.1 & 46.6 & 10.8 & 12.9 & 23.5 \\
Diffusion\cite{li2023grounded} & \cmark & \textbf{67.4} & \textbf{27.1} & \textbf{43.6} & \textbf{45.9} & \textbf{63.6} & \textbf{48.8} & \textbf{42.0} & \textbf{38.6} & \textbf{48.9} & \textbf{11.3} & \textbf{44.2} & \textbf{34.9} \\
\hline
\multirow{2}{*}{OVAM} & \xmark & 49.9 & 31.4 & 28.0 & 25.9 & 51.2 & 54.0 & 15.2 & 23.5 & 42.6 & 10.9 & 38.2 & 30.0 \\
                      & \cmark & \textbf{57.5} & \textbf{32.2} & \textbf{44.6} & \textbf{41.1} & \textbf{58.1} & \textbf{55.2} & \textbf{42.4} & \textbf{28.0} & \textbf{44.4} & \textbf{22.4} & \textbf{51.9} & \textbf{36.1} \\
\hline
\end{tabular}
\caption{Evaluation of VOC challenge performance for a model trained on synthetic data, comparing the impact of token optimization.}
\label{tab:additional-synthetic-training-comparison}
\end{table}

\subsection{Presence of token in prompts}
\label{subsect:additional-presence-tokens}

To explore the impact of explicitly mentioning the word used for extracting attentions (attribution prompt) within the image synthesis prompt (generator prompt), Table \ref{tab:token-presence} expands on the overview provided in Table 1 (Section 4.1). This table breaks down the COCO-cap results by class and distinguishes between cases where the class name—used for mask generation—is included in the generator prompt or not. This detailed evaluation reveals no discernible trend to suggest that the explicit inclusion of the token in the prompt markedly influences the mIoU of the generated masks.

\begin{table}[htpb]
\centering
\setlength{\tabcolsep}{3pt}
\begin{tabular}{c|c|ccccccccccc|c}
\hline
\multirow{2}{*}{Method} & \multirow{2}{*}{\shortstack{Token\\included}} & \multicolumn{11}{c|}{Selected classes (COCO-cap IoU \%)} & \multirow{2}{*}{mIoU} \\
\cline{3-13}
       &       & aeroplane & bicycle & bird & boat & bus & car & cat & dog & horse & person & train &  \\
\hline
\multirow{3}{*}{DAAM \cite{DAAM}} & \xmark & 37.3 & 33.0 & 47.9 & 26.5 & 77.6 & 54.0 & 86.2 & 80.8 & 51.5 & 23.9 & 55.9 & 48.1 \\
                      & \cmark & 24.7 & 34.3 & 56.8 & 33.3 & 84.3 & 38.8 & 80.5 & 76.5 & 49.1 & 21.6 & 40.4 & 48.7 \\ \cdashline{2-14}
                      & \emph{all} & 30.6 & 33.8 & 53.0 & 31.9 & 82.6 & 42.8 & 83.0 & 77.9 & 49.8 & 22.7 & 44.6 & 48.4 \\
\hline
\multirow{3}{*}{DatasetDM\cite{DatasetDM}} & \xmark & 75.8 & 19.4 & 85.1 & 78.1 & 80.2 & 37.2 & 83.2 & 74.9 & 82.0 & 56.7 & 51.6 & 60.2 \\
                           & \cmark & 72.3 & 29.6 & 89.0 & 69.7 & 95.9 & 59.3 & 72.1 & 68.8 & 83.9 & 48.2 & 85.2 & 58.9 \\ \cdashline{2-14}
                           & \emph{all} & 74.1 & 25.7 & 87.4 & 71.3 & 91.4 & 51.9 & 76.2 & 71.7 & 83.5 & 52.2 & 72.9 & 59.3 \\
\hline
Grounded  & \xmark & 84.3 & 47.9 & 80.0 & 61.5 & 94.9 &  0.0 & 89.6 & 83.1 & 86.6 & 53.7 & 67.9 & 52.0 \\
Diffusion & \cmark & 85.0 & 58.4 & 83.4 & 17.7 & 80.5 & 27.4 & 85.9 & 85.3 & 83.7 & 49.2 & 44.5 & 47.9 \\ \cdashline{2-14}
\cite{li2023grounded} & \emph{all} & 84.6 & 54.9 & 81.9 & 30.8 & 81.4 & 25.1 & 87.3 & 84.4 & 84.1 & 51.8 & 47.2 & 50.2 \\
\hline
\multirow{3}{*}{OVAM} & \xmark & 77.8 & 67.5 & 52.4 & 46.0 & 83.8 & 45.5 & 70.5 & 64.1 & 66.1 & 25.0 & 58.2 & 58.4 \\
                      & \cmark & 53.6 & 62.7 & 56.2 & 53.4 & 85.2 & 48.5 & 65.8 & 66.7 & 74.4 & 15.2 & 50.9 & 58.3 \\ \cdashline{2-14}
                      & \emph{all} & 65.1 & 64.3 & 54.6 & 51.9 & 84.9 & 47.5 & 67.9 & 65.8 & 71.5 & 19.7 & 53.0 & 58.2 \\
\hline
\end{tabular}
\caption{Table comparing mIoU whether word used for pseudo-mask generation is included in the generator prompt.}
\label{tab:token-presence}
\end{table}

\clearpage
\section{Qualitative Examples}
\label{sec:B}
\subsection{Qualitative comparison of OVAM Attention Maps}

\clearpage
\subsection{Use of OVAM-optimized Tokens with Other Methods}

\clearpage

\end{document}